\documentclass[lettersize,journal]{IEEEtran}
\usepackage{amsmath,amsfonts}
\usepackage{algorithm}
\usepackage{array}
\usepackage[caption=false,font=normalsize,labelfont=sf,textfont=sf]{subfig}
\usepackage{textcomp}
\usepackage{stfloats}
\usepackage{url}
\usepackage{verbatim}
\usepackage{graphicx}
\usepackage{cite}
\usepackage{amsfonts,amssymb}
\usepackage{bm}
\usepackage{multicol}
\usepackage{booktabs}
\usepackage{bbding}
\usepackage{footnote,threeparttable}
\usepackage[english]{babel}
\usepackage{blindtext}
\usepackage{color}
\usepackage{algpseudocode}
\usepackage{tikz}
\usepackage{textcomp}
\usepackage{hyperref}
\usepackage{lipsum}

\begin{document}

%\title{The Design and Control of a Manipulator-Assisted Landing System on a Disturbed USV}

% \onecolumn 
%{\huge{\textbf{IEEE Copyright Notice}}
%	\\
%
%\large{This work has been submitted to the IEEE for possible publication. Copyright may be transferred without notice, after which this version may no longer be accessible. DOI: https://doi.org/10.48550/arXiv.2212.12196
%	
%	Copyright (c) 2023 IEEE
%	Personal use of this material is permitted. Permission from IEEE must be obtained for all other uses, in any current or future media, including reprinting/republishing this material for advertising or promotional purposes, creating new collective works, for resale or redistribution to servers or lists, or reuse of any	copyrighted component of this work in other works.}}
%\twocolumn 

 \onecolumn 
{\huge{\textbf{Copyright Notice}}
	\\
	
	\large{This work has been submitted to the Elsevier for possible publication. Copyright may be transferred without notice, after which this version may no longer be accessible.
		
	}}
\twocolumn 

\title{A Manipulator-Assisted Multiple UAV Landing System for USV Subject to Disturbance}

%\author{IEEE Publication Technology,~\IEEEmembership{Staff,~IEEE,}
\author{\IEEEauthorblockN{Ruoyu Xu, Chongfeng Liu, Zhongzhong Cao, Yuquan Wang and Huihuan Qian}

        % <-this % stops a space

\thanks{Ruoyu Xu, Chongfeng Liu, Zhongzhong Cao and Huihuan Qian are with Shenzhen Institute of Artificial Intelligence and Robotics for Society (AIRS), The Chinese University of Hong Kong, Shenzhen (CUHK-Shenzhen), Shenzhen, Guangdong, 518129, China.}
\thanks{Ruoyu Xu, Chongfeng Liu, and Huihuan Qian are also with School of Science and Engineering (SSE), The Chinese University of Hong Kong, Shenzhen (CUHK-Shenzhen), Shenzhen, Guangdong 518172, China. }
\thanks{Yuquan Wang is with the 
	Department of Advanced Computing Sciences, Maastricht University, Maastricht, the Netherlands.}
\thanks{Corresponding author: Huihuan Qian (email: hhqian@cuhk.edu.cn).}

}

% The paper headers
%\markboth{ IEEE Transactions on Systems, Man, and Cybernetics: Systems}%

\markboth{ }
{Shell \MakeLowercase{\textit{et al.}}: A Sample Article Using IEEEtran.cls for IEEE Journals}

%\IEEEpubid{0000--0000/00\$00.00~\copyright~2021 IEEE}
% Remember, if you use this you must call \IEEEpubidadjcol in the second
% column for its text to clear the IEEEpubid mark.

\maketitle
%\copyrightnotice
\begin{abstract}

Marine waves significantly disturb the unmanned surface vehicle (USV) motion. An unmanned aerial vehicle (UAV) can hardly land on a USV that undergoes irregular motion. An oversized landing platform is usually necessary to guarantee the landing safety, which limits the number of UAVs that can be carried. We propose a landing system assisted by tether and robot manipulation. The system can land multiple UAVs without increasing the USV's size. An MPC controller stabilizes the end-effector and tracks the UAVs, and an adaptive estimator addresses the disturbance caused by the base motion. The working strategy of the system is designed to plan the motion of each device. We have validated the manipulator controller through simulations and well-controlled indoor experiments. During the field tests, the proposed system caught and placed the UAVs when the disturbed USV roll range was approximately 12 degrees.

\end{abstract}

\begin{IEEEkeywords}
landing assistance system, UAV recovery, motion control, adaptive estimator.
\end{IEEEkeywords}

\section{Introduction}

Cooperative unmanned surface vehicle (USV) and unmanned aerial vehicle (UAV) systems are increasingly important in tasks such as coastal patrol, surface rescue, environmental exploration, and monitoring \cite{zhang2020marine}. The UAV has a large field of view and higher mobility. However, its limited energy onboard results in the necessary landing onto a USV for recharging.

It is essential and challenging to land UAVs on a USV subject to wave disturbances in the sea. The maximum pitch angle of a two-meter USV under level three sea condition is approximately 8 degrees \cite{lv2021analysis}, and the roll angle of a six-meter USV is usually more than 5 degrees when working under high sea state \cite{peng2017development}. The disturbed landing platform leads to an unexpected ground effect \cite{matus2021ground} that reduces the landing accuracy. Besides, the shaking and heaving of the USV can collide with the UAV \cite{maier2020coordinated}. Although there are cases of UAVs successfully landing on USVs, a large landing platform is usually applied to guarantee landing safety \cite{xu2020vision,ambroziak2022experimental,borreguero2018experimental}. Therefore, most traditional USV-UAV systems only carry one UAV due to the limitation of the landing area. 
\begin{figure}[t]
	\centering
	\setlength{\abovecaptionskip}{0pt}
	\setlength{\belowcaptionskip}{0pt}
	\includegraphics[width=1\linewidth]{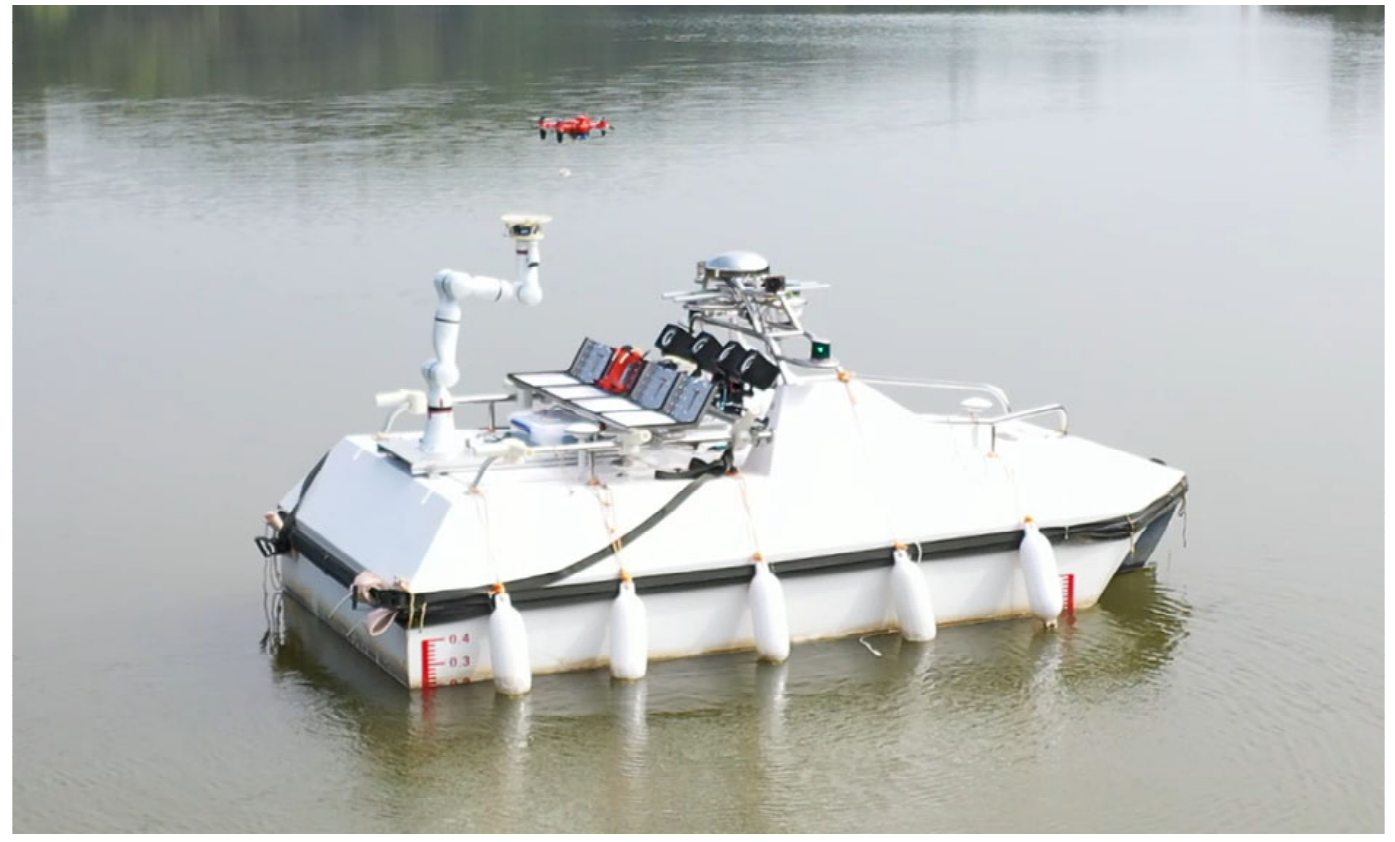}
	\caption{Illustration of the landing assistance system.}
	\label{demo}
\end{figure}

Using assistance systems can efficiently improve landing accuracy and safety. Although it is difficult to land the UAV directly, it is easier to control the UAV and maintain its relative position to the USV. Thereafter, it is possible for a manipulator to catch the UAV precisely, and the demand for a large landing platform is eliminated, which makes the USV possible to carry multiple UAVs. However, catching the UAV directly in a disturbed marine environment is still venturesome. Tethered landing can avoid direct collision, but it is difficult to establish a tethered connection. Therefore, we pursue a strategy to combine the advantages of tethered landing and manipulator-assisted landing.

In this paper, a manipulator-based assistance system for UAV landing on USV is proposed. As shown in Fig. \ref{demo}, the system consists of a seven-DoF manipulator with a catcher and a tethered landing system on the UAV. Four landing platforms are designed capable of locking and releasing the UAVs. An infrared camera system is employed for the UAV position feedback. A tethered landing system is designed for the UAV to accurately land on the catcher, and the manipulator can track and establish the tethered connection with the UAV. After the UAV is fastened to the catcher, the manipulator can place it onto the landing platform accurately. 

From the perspective of control, accurate capturing of UAV by the manipulator is challenging and affected by the wave disturbance. Unexpected base motion brings time-varying dynamics and disturbance, which degrades the performance of the manipulator controller. In addition, joint constraints need to be considered, especially in continuous tracking tasks. An MPC-based controller can take joint state constraints into consideration. The controller treats the wave-induced disturbance as uncertainty, and an adaptive estimator is designed for feedforward compensation.

The contributions of this paper are summaries as follows:
\begin{itemize}
	\item We designed and studied a manipulator-assisted UAV landing system for the USV. The system is capable of landing UAVs on the USV with limited deck space accurately and safely, making the USV possible to carry multiple UAVs. Besides, the tethered connection between the UAV and manipulator can significantly increase the landing efficiency.
	
	\item To improve the accuracy of the manipulator on the USV, we overcome the disturbance caused by base motions as uncertainty in joint space. An MPC controller with an uncertainty estimator is designed for objective tracking and catching on a disturbed USV. 
	
	\item A series of experiments is conducted to verify the proposed manipulator controller, and field experiments are conducted to test the functionality of the landing assistance system.
\end{itemize}

The remainder of this paper is organized as follows. The related work is reviewed and given in Section \ref{review}. Section \ref{design} introduces the system structure and the function of every subsystem of the landing assistance system. Section \ref{control} elaborates the manipulator controller design and the working strategy of the landing system. Both simulations and experiments are conducted in Section \ref{exp} to verify the proposed controller, and the functionality of the system is tested by field experiments.

\section{Related Work}
\label{review}
\subsection{UAV Landing Method}
A UAV can land accurately on a landing platform without disturbance by visual servo \cite{lin2021low}. Multiple scales of marks are used for the positioning of the landing platform from different distances \cite{grlj2022decade}. For the case of landing on a moving platform, UAV can predict the target motion using the extended Kalman filter \cite{araar2017vision} or the autoregressive model \cite{huang2021linear}. The landing algorithm for a perpetuated platform subject to different sea conditions is also investigated \cite{polvara2018vision}, but its performance is not verified by practical experiments. Model predictive control (MPC) can also be applied to find an optimal control input for autonomous landing \cite{guo2022autonomous}. The robust controller can be applied to achieve fast and precise landing under the effect of winds \cite{ paris2020dynamic}. In summary, existing algorithms can realize accurate landing for a slowly moving target or a platform performing regular movement \cite{baca2019autonomous}. The motion of the USV subject to wave disturbance is fast and stochastic, so the high-precise landing is still a challenge in this condition.

\subsection{Landing Assistance Method}
It is natural to utilize a landing assistance device if the pure algorithm cannot guarantee high precision. USVs can use an independent suspension system to stabilize the deck \cite{sinisterra2017usv} or apply a Stewart platform to balance the USV motion and provide a stable landing platform \cite{zhang2020marine,ross2022autonomous}. However, the motion range of these systems is limited, and the ground effect exists, which brings disturbed airflow and thus makes it difficult for precise landing. The tethered system \cite{oh2006approaches} establishes a connection between the USV and the UAV, and the UAV can be recycled by winding up the string on the USV. This method can resist wind disturbance and realize fixed-point landing \cite{kourani2021marine}, but the string length limits the motion range of the UAV \cite{kishk20203,liu2022design}, and strings will be entangled in the multi-UAV task \cite{kominami2021active}. Some tethered UAVs are designed to be detachable, and the tether can attach to the deck by electromagnetic winch \cite{schuchardt2021maritime}, or compliant grapple \cite{nguyen2019passively}. But to the best of our knowledge, there is no tethered UAV that can reconnect to a specific point autonomously. The other method is using the manipulator to assist the UAV landing \cite{maier2015robot}. Manipulator has a larger range of motion compared with the Stewart platform, so it is capable of tracking the UAV and balancing the USV motion at the same time. However, existing researches mainly focus on the touchdown of the UAV, and the UAV is docked to the manipulator end-effector aforehand. Establishing the connection between the UAV and the end-effector on a perpetuated platform is still a problem.

In this paper, the tethered landing method and the manipulator-assisted landing method are combined. The tethered connection is established by using the manipulator to catch the suspension released by the UAV. The UAV is accurately placed on the platform after it is captured by tethered landing.

\subsection{Controller for Floating-Base Manipulator}
Existing research modeled the dynamics of the floating manipulator on the surface \cite{from2010singularity,from2011motion}, and the coupling between the manipulator and the base motion can be considered in the controller \cite{kosuge1991control}. The model of floating manipulator with two joints can be easily acquired \cite{xu2020novel}. However, the full dynamics of the floating manipulator with high degrees of freedom (DoF) is very complex, especially for dexterous manipulators. Robust controllers like sliding mode control \cite{yuguang2019dynamic,dai2022fast} and backstepping control \cite{zhang2021manipulator} are other methods for addressing the uncertainty, but complex design is usually involved \cite{guerrero2019observation}. The MPC controller can also be applied in this condition \cite{qiang2020model}. An autoregressive model  \cite{woolfrey2021predictive} or wavelet network \cite{xu2021predictive} is applied to generate the trajectory in the prediction horizon, and then the optimal solution can be implemented for better tracking performance. However, These MPC controllers are formulated based on a nominal model, and the wave-induced motion brings disturbances, which degrades the tracking accuracy. The disturbance can be estimated by the adaptive neural network \cite{qian2021adaptive} or the observer \cite{han2021adaptive,xu2022design}. Therefore, an uncertainty estimator is designed in this paper, serving as the feedforward controller of the MPC tracking controller.

\section{System Design}
\label{design}
\begin{figure}[t]
	\centering
	\setlength{\abovecaptionskip}{0pt}
	\setlength{\belowcaptionskip}{0pt}
	\includegraphics[width=1\linewidth]{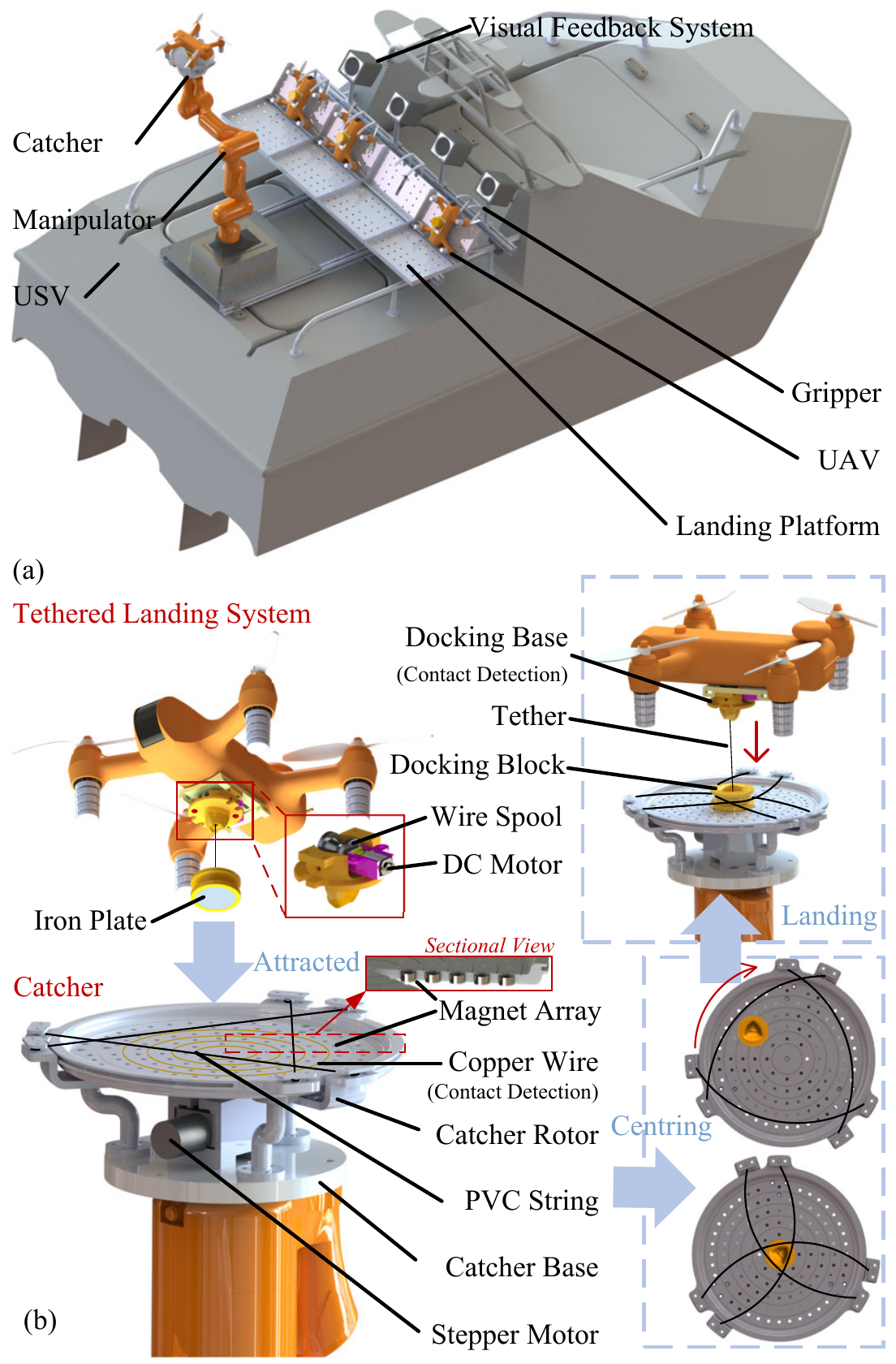}
	\caption{The assistance landing system based on the manipulator. (a) is the overall concept of the system on a USV. (b) shows the detailed structure of the tethered landing system and the catcher, and the tethered landing process.}
	\label{overall}
\end{figure}

\begin{figure}[t]
	\centering
	\setlength{\abovecaptionskip}{0pt}
	\setlength{\belowcaptionskip}{0pt}
	\includegraphics[width=1\linewidth]{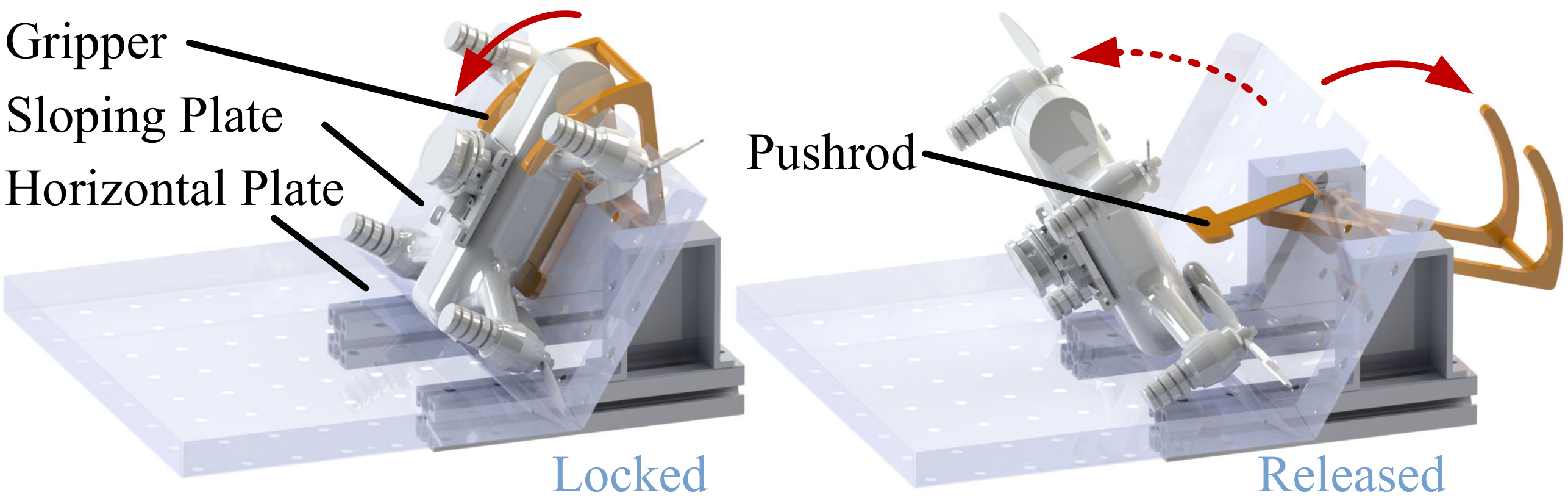}
	\caption{The UAV locking and releasing on the landing platform.}
	\label{plate}
\end{figure}

\begin{figure}[t]
	\centering
	\setlength{\abovecaptionskip}{0pt}
	\setlength{\belowcaptionskip}{0pt}
	\includegraphics[width=1\linewidth]{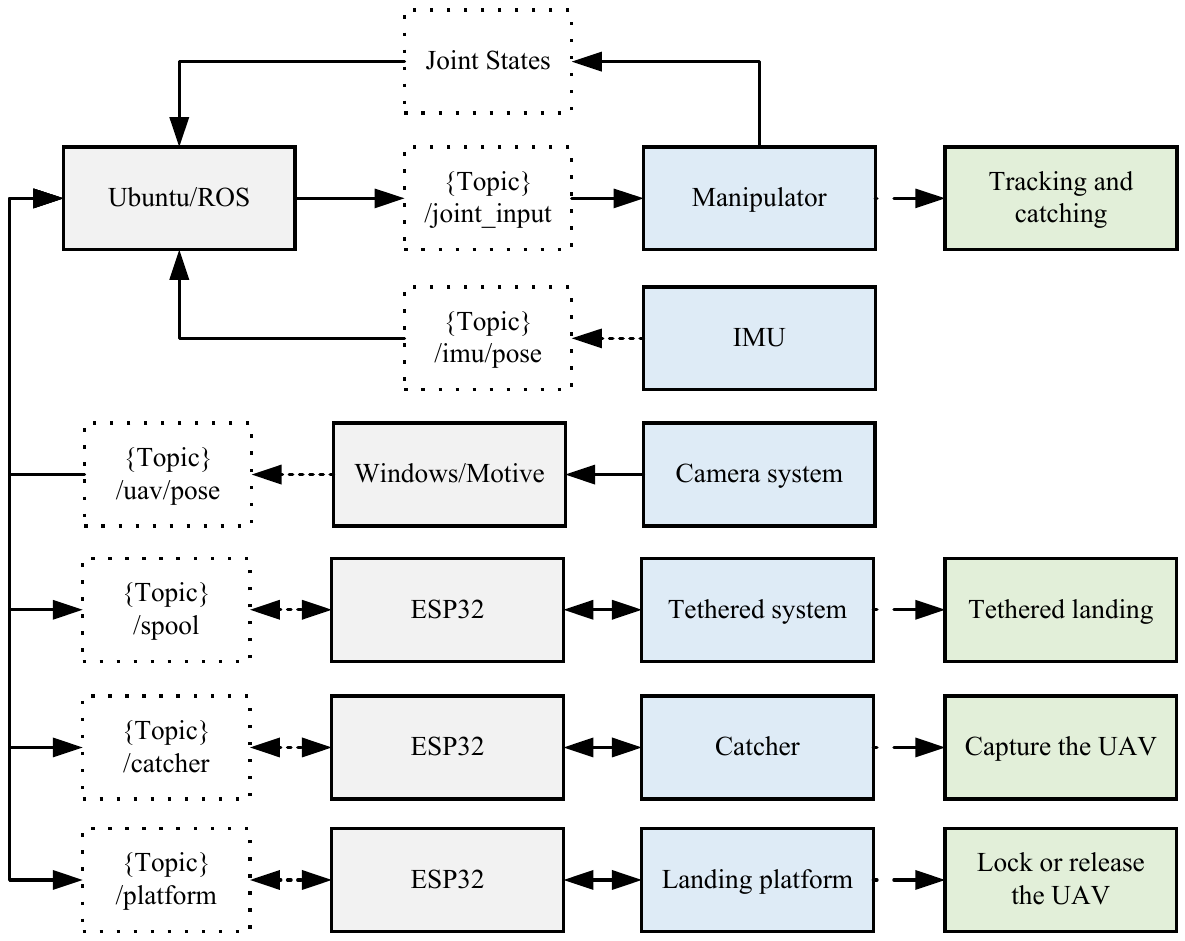}
	\caption{Schematic of the system. The gray boxes are the controllers in this system, the blue boxes show the subsystems, and the green boxes show the function of each subsystem. The subsystems are communicated through ROS topic, which is shown in dotted box.}
	\label{comm}
\end{figure}

This section presents the overall concept of our system design. As shown in Fig. \ref{overall}(a), the manipulator is mounted on the USV to track and catch the UAV. A catcher is developed as the end-effector of the manipulator. A tethered landing system mounted on UAV is designed to land the UAV on the catcher accurately and safely. A landing platform with gripper is also designed to fix the UAV on the USV,.

The landing process is shown in Fig. \ref{overall}(b), the UAV releases the docking block with an iron plate and hovers in the dexterous space of the manipulator. The manipulator catches the docking block, and the catcher will lock the block in a second. The docking structure reels the tether to land the UAV on the catcher. Thereafter, the manipulator will place the UAV on the landing platform, and the gripper on the landing platform will keep the UAV fixed. In the landing process, the position of the UAV is captured in real-time by the visual feedback system. 

Compared with landing the UAV on a disturbed platform directly, it is easier for the UAV to hover above the USV. Using a manipulator to recycle the UAV improves the landing accuracy. The tethered landing system avoids direct contact between the UAV and the fluctuating landing platform, and also reduces the effect of the ground effect. The details of this system will be elaborated in the remainder of this section.

\subsection{Manipulator and Catcher}
\label{catcher}
A redundant manipulator is utilized in this system, which brings larger dexterous space and better tolerance to joint state saturation. The manipulator is capable of balance the base motion and catches the hovering UAV, which reduces the counterblast. In this way, the ground effect during the landing process is avoid.

The catcher is designed to (1) fix the UAV quickly and (2) center the UAV for accurate placement. As shown in Fig. \ref{overall}(b), the catcher consists of a stepper motor, catcher base, and catcher rotor. A magnet array is embedded in the bottom of the catcher base, as shown in the sectional view, to attract the docking block of the UAV quickly. Three PVC strings are utilized to lock the docking block to prevent the connection from being broken up by the unexpected base motion. As shown in Fig. \ref{overall}(b), both the rotor and base have three clampers, one side of the string is fixed to the base, and the other is connected to the rotor. Rotating the rotor can shrink the vacant space within three strings and push the docking block to the center of the catcher. A stepper motor with a reducer drives the rotor, and the strings can be tightened within 0.8 seconds. Because the docking block is designed to have a triangle surface contacted with three strings, the rotation is constrained after the catcher is locked.

To lock the docking block at the moment it is attracted, copper wires are embedded into the surface of the catcher for contact detection. Copper wires are arranged in concentric circles, and the distance between two wires is less than the diameter of the docking block. The adjacent wires will be connected at the moment of the docking block contact the surface of the catcher. One of the adjacent wires is connected to the ground, and the other is connected to the GPIO of the controller on the catcher. The GPIO is connected to the 3.3 V power source through a pull-up resistor. Before the docking block is attracted, the GPIO voltage is high. An external interrupt is triggered by the low voltage to lock the docking block as long as the contact is detected.

\subsection{Tethered Landing System}
To accurately land the UAV on the catcher and avoid the collision between catcher and UAV, a tethered landing system shown in Fig. \ref{overall}(b) is designed and mounted on the bottom of the UAV. The structure consists of a docking block, a docking base, and a wire spool. The bottom of the docking block is an iron plate, and the top of the block has a concave whose shape is a Reuleaux tetrahedron. The docking block is connected to the docking base by a tether and can be released by a wire spool controlled by a DC motor. The manipulator can catch the block without having direct contact with the UAV, and the UAV can land on the manipulator by reeling the wire spool. %This landing process is the same as these tethered UAVs, so the advantages of tethered UAVs are retained after the connection between the docking block and manipulator is established.

The bottom of the docking base is a raised Reuleaux tetrahedron and can be docked into the concave of the docking block. This docking process can be easily realized under the effect of tether tension. As long as the docking is finished, the rotation and position of the UAV will be constrained by the tetrahedral docking structure. The tethered landing system also has a contact detection module under the docking base, which works the same way as the contact detection module on the catcher. The success of the docking process will also generate a signal to the controller of the tethered landing system.

\subsection{Landing Platform}
As shown in Fig. \ref{plate}, the landing platform is designed to fix the UAV on the USV, consisting of a sloping plate and a horizontal plate. A gripper driven by a steering motor is mounted behind the sloping platform. After the UAV is placed on the sloping plate, the gripper can rotate downward to fix the UAV so as to prevent the UAV from dropping out of the platform. A pushrod can pull the UAV down to the horizontal plate for takeoff by rotating the gripper to the other side.  

Because the manipulator can place the UAV accurately, the size of the Landing platform does not need to be very large to guarantee landing safety. Therefore, multiple UAVs can be stored on one USV. In our prototype system, one USV is capable of carrying four UAVs.

\subsection{Visual Feedback System}
A motion capture subsystem with four infrared cameras (i.e., Optitrack Prime41) is utilized to perceive the position of the UAV. Although the monocular or stereo camera is common in similar visual feedback systems, their feedback frequency is usually limited by the performance of the controller. By using motion capture technology based on infrared cameras, the feedback frequency reaches 200 Hz, which is sufficient for UAV catching. Besides, this visual feedback system can work well in a dark environment, which makes the UAV can be recycled in rayless surroundings.

Four cameras are evenly arrayed on the USV, and the interval distance is 0.37 m. Markers are mounted on the UAV for localization. Because the UAV needs to be exposed to at least two cameras to acquire the position, the poses of the cameras are adjusted to ensure the dexterous space of the manipulator can be covered.

\subsection{Control System}
The control system of the landing system is constructed based on Robot Operation System (ROS), as shown in Fig. \ref{comm}. A wireless station is applied to connect all the devices on USV to a local area network. The landing platform, docking structure, and the catcher are controlled by three microcontrollers, respectively. The microcontroller we applied is ESP32 equipped with a wireless module. The infrared cameras are connected to a computer with a Windows system by a gateway that supports the Power Over Ethernet (POE) protocol. A computer with an Ubuntu system controls the manipulator. All the microcontrollers and computers are connected to the onboard wireless station and share the same network segment. ROS is deployed on the computer with the Ubuntu system, which is the host controller. In this way, the controller of these devices can communicate in real-time, and each controller is a node in the ROS. This centralized control method benefits the management of each device. Besides, external sensors like IMU and cameras can be easily fused into the control system by this structure.

\section{Controller Design}
\label{control}
\begin{figure}[t]
	\centering
	\setlength{\abovecaptionskip}{0pt}
	\setlength{\belowcaptionskip}{0pt}
	\includegraphics[width=1\linewidth]{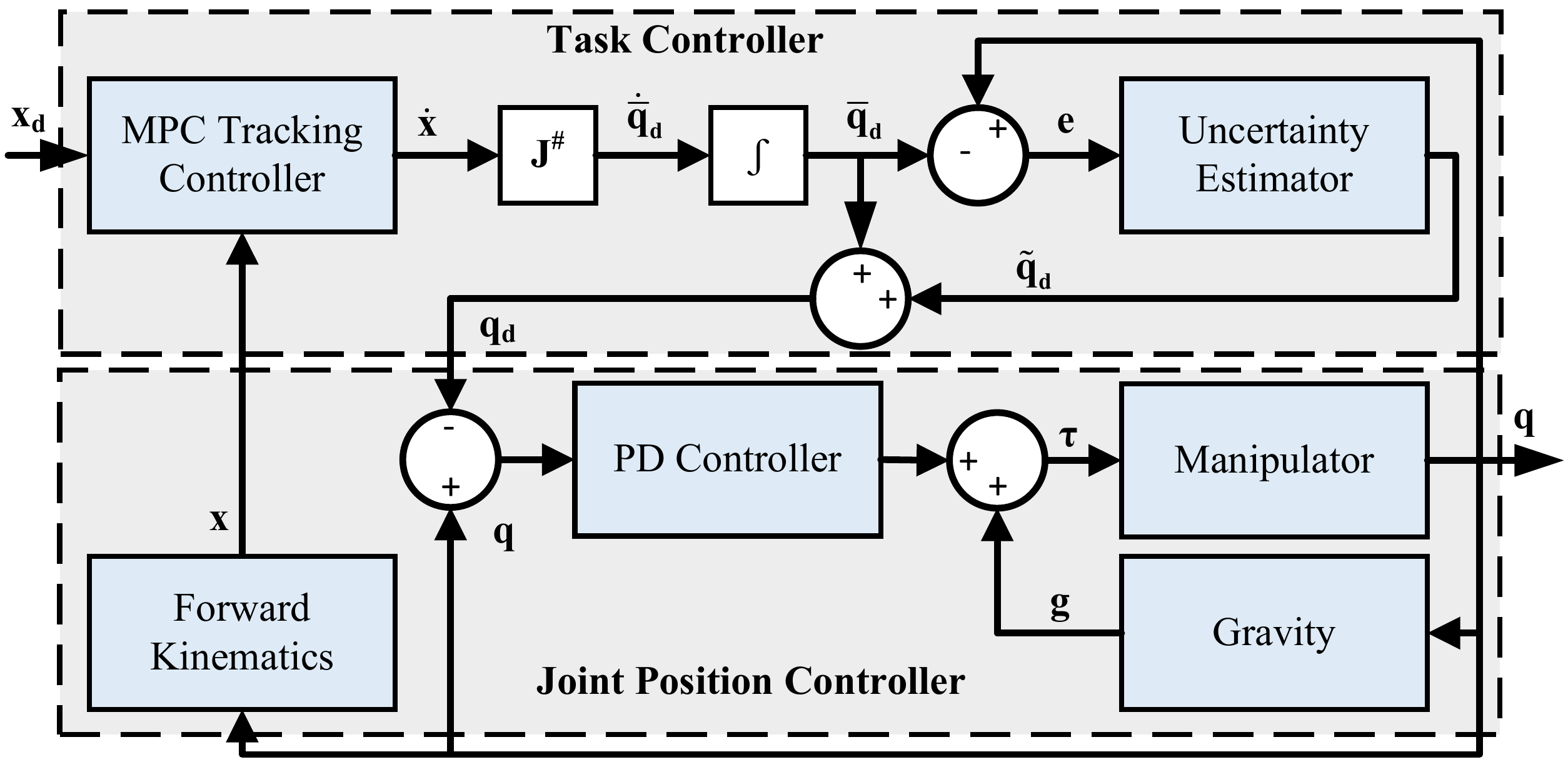}
	\caption{The control diagram of the manipulator.}
	\label{controller}
\end{figure}

This section introduces the controller design of the landing assistance system. We start from the floating manipulator controller, and the control diagram is given in Fig. \ref{controller}. The controller consists of a task controller and a joint position controller. The joint position controller is a PD controller with gravity compensation. In the task controller, the uncertainty estimator is designed to improve the accuracy under the effect of base motion, which is introduced in Section \ref{UEFM}, and the tracking controller based on MPC is introduced in Section \ref{OTC}. A Kalman filter is designed to improve the position feedback performance of the camera system, which is introduced in  Section \ref{UPF}. At last, the control strategy for landing assistance is given in Section \ref{SCS}.
\subsection{Uncertainty Estimator for Floating Manipulator}
\label{UEFM}
%The manipulator is the main objective to be considered in the controller design. There are two problems for the manipulator mounted on a shaking USV. The first problem is the UAV tracking problem under the effect of base motions. Differing from the fixed-base manipulator, unknown dynamics caused by base motion can degrade the tracking performance of the manipulator. The other problem is the smooth control problem, especially when the joint position is close to its limitation. Some commercial manipulators have protection mechanism when the manipulator reaches the joint constraints, and hard stop can be triggered in this condition, which could damage the joint if the manipulator is running in high speed. To solve these problems, a hierarchical control framework for this redundant manipulator is applied. The low-level controller is used to control the joint positions, and the high-level controller is designed in the Cartesian space.
%The dynamic model of the floating manipulator has the form as \eqref{eq3_1},
An uncertainty estimator is designed for the floating manipulator to reduce the joint tracking error induced by base motion. The proposed estimator is formulated as an added joint position command. Differing from most uncertainty estimator that functions as a torque feedforward controller, the proposed estimator can be applied to a manipulator with only a position control interface. Besides, the uncertainty estimator is formulated without the knowledge of the complex dynamic model of a floating redundancy manipulator. 

The motion of the manipulator can be affected by the inertial torque induced by base motion, joint motion, and gravity. Consider a floating manipulator whose dynamic model has the form as \eqref{eq3_1},
\begin{equation}
	{{\bf{M}}}{{{\bf{\ddot q}}}} + {{\bf{C}}}{{{\bf{\dot q}}}} + {{\bf{M}}_{\rm{mu}}}{{{\bf{\ddot q}}}_{\rm{u}}} + {{\bf{C}}_{\rm{mu}}}{{{\bf{\dot q}}}_{\rm{u}}} + {{\bf{g}}} = {{\bm{\tau }}}
	\label{eq3_1}
\end{equation}
where ${\bm{\tau }}$ is the joint torque input, ${\bf{q}}$ and ${{\bf{q}}_{\rm{u}}}$ are the joint position and base motion, ${\bf{M}}$ and ${\bf{C}}$ are the matrices related to the inertial and Coriolis forces, ${\bf{M}}_{\rm{mu}}$ and ${\bf{C}}_{\rm{mu}}$ are related to the inertial and Coriolis forces caused by the base motion, ${\bf{g }}$ represents gravity. The joint position can be controlled by the PD controller with complete dynamics compensation. The gravity term usually has analytic expression, and the torque induced by the base motion and the joint velocity is expressed by \eqref{eq3_3}.
\begin{equation}
	{{\bm{\tau }}_{\rm{e}}} = {\bf{M\ddot q}} + {\bf{C\dot q}} + {{\bf{M}}_{\rm{mu}}}{{{\bf{\ddot q}}}_{\rm{u}}} + {{\bf{C}}_{\rm{mu}}}{{{\bf{\dot q}}}_{\rm{u}}}
	\label{eq3_3}
\end{equation}

The real-time computation of \eqref{eq3_3} brings a heavy computational load. Therefore, it is common for commercial manipulators to use a PD controller with gravity compensation for the joint position control, as given in \eqref{eq3_2},
\begin{equation}
	{{\bm{\tau }}} =  - {{\bf{K}}_{\rm{p}}}\left( {{{\bf{q}}} - {{\bf{q}}_{\rm{d}}}} \right) - {{\bf{K}}_{\rm{d}}}\left( {{{{\bf{\dot q}}}} - {{{\bf{\dot q}}}_{\rm{d}}}} \right) + {{\bf{g}}}
	\label{eq3_2}
\end{equation}
where ${{\bf{q}}_{\rm{d}}}$ is the desired joint position, $ {{\bf{K}}_{\rm{p}}}$ and ${{\bf{K}}_{\rm{d}}}$ are positive-definite diagonal gain matrices. This method avoids the requirement for accurate modeling and complex computation, but the control accuracy is less than the computed torque control with complete dynamics compensation \cite{leahy1989compensation}, especially in the task of tracking a moving object. The torque control, like the iterative Newton-Euler method, can consider the inertia and Coriolis forces. However, torque control is not available for some commercial manipulators, and the computation is still complex if the effect of base motion is considered.

To compensate the torques related to inertia and velocity, an adaptive estimator is designed. The estimator is applied as an added joint position command ${{\bf{\tilde q}}_{\rm{d}}}$, and it estimates the joint position errors caused by base motion. Let the actual desired joint position be ${{\mathbf{\bar q}}_{\rm{d}}}$, so the new position command in \eqref{eq3_2} can be written as \eqref{eq3_4a}.
\begin{equation}
	{{\bf{q}}_{\rm{d}}} = {{\mathbf{\bar q}}_{\rm{d}}}+{{\bf{\tilde q}}_{\rm{d}}} 
	\label{eq3_4a}
\end{equation}
Let the joint position error be ${\bf{e}} = {\bf{q}} - {{\mathbf{\bar q}}_{\rm{d}}}$, and the desired joint velocity be zero. Thereafter, the joint dynamics after the estimator is applied can be derived from \eqref{eq3_2} and \eqref{eq3_4a}, and the dynamics for one joint is \eqref{eq3_4}.
\begin{equation}
	{k_{{\rm{p,}}i}}{e_i} + {k_{{\rm{d,}}i}}{\dot e_i} = {k_{{\rm{p,}}i}}{\tilde q_{{\rm{d,}}i}} - {{{\tau }}_{{\rm{e,}}i}},i = 1,...,7
	\label{eq3_4}
\end{equation}
${k_{{\rm{p,}}i}}$ and ${k_{{\rm{d,}}i}}$ are the parameters in ${{\bf{K}}_{\rm{p}}}$ and ${{\bf{K}}_{\rm{d}}}$ that related to the joint $i$. ${e_i}$, ${\tilde q_{{\rm{d,}}i}}$, and ${{{\tau }}_{{\rm{e,}}i}}$ are the $i$-th component of ${\bf{e}}$, ${{\bf{\tilde q}}_{\rm{d}}}$, and  ${{\bm{\tau }}_{{\rm{e}}}}$. The unknown dynamics of each joint is a function of errors, and it can be linearized by Taylor expansion as the method in \cite{ahmadi2016robust}. Let the approximation error of one joint be be $\varepsilon_{e,i}$, and the error is bounded by a positive value $\mu$, $|\varepsilon_{e,i}|<\mu$. Thereafter, the linearized result is shown in \eqref{eq3_7}.
\begin{equation}
	{\tau _{e,i}} = {\tau _{0,i}} + \frac{{\partial {\tau _{e,i}}}}{{\partial {e_i}}}{{\rm{e}}_i}{\rm{ + }}\frac{{\partial {\tau _{e,i}}}}{{\partial {{{\rm{\dot e}}}_i}}}{{\rm{\dot e}}_i} + \varepsilon_{e,i}
	\label{eq3_7}
\end{equation}
Define ${\tau _{0,i}}$, $\partial {\tau _{e,i}}/\partial {e_i}$, and $\partial {\tau _{e,i}}/\partial {{{{\dot e}}}_i}$ as ${\lambda _1}$, ${\lambda _2}$, and ${\lambda _3}$. The equation \eqref{eq3_7} can be further rewritten as \eqref{eq3_8},
\begin{equation}
	{\tau _{{\rm{e,}}i}} = {{\bm{\lambda }}^{\rm{T}}}{\bm{\delta }} + \varepsilon_{e,i}
	\label{eq3_8}
\end{equation}
where ${\bm{\lambda}} = \left[{{\lambda _1}},{{\lambda _2}},{{\lambda _3}}\right]^{\rm{T}}$ and ${\bm{\delta }} = {\left[
			1,{{e_i}},{{{\dot e}_i}}\right]^{\rm{T}}}$. Let the estimator have the form as \eqref{eq3_8b},
\begin{equation}
	{\tilde q_{{\rm{d,}}i}} = {{\bm{\hat \lambda }}^{\rm{T}}}{\bm{\delta }}/{k_{{\rm{p}},i}}
	\label{eq3_8b}
\end{equation}
where ${\bm{\hat \lambda }} = {\left[{{{\hat \lambda }_{1}}},{{{\hat \lambda }_{2}}},{{{\hat \lambda }_{3}}} \right]^{\rm{T}}}$ is the estimation of parameter vector ${\bm{\lambda }}$ in \eqref{eq3_8}. After \eqref{eq3_8b} is applied, the dynamics \eqref{eq3_4} can be rewritten as \eqref{eq3_9}.
\begin{equation}
	{k_{{\rm{p}},i}}{e_i} + {k_{{\rm{d}},i}}{\dot e_i} = \left( {{{{\bm{\hat \lambda }}}^{\rm{T}}} - {{\bm{\lambda }}^{\rm{T}}}} \right){\bm{\delta }} -\varepsilon_{e,i}
	\label{eq3_9}
\end{equation}
The dynamics of each joint can be derived from \eqref{eq3_9}, and rewritten as \eqref{eq3_10}.
\begin{equation}
	{\dot e_i} =  - {k_{{\rm{p,}}i}}/{k_{{\rm{d,}}i}}{e_i} + \left( {{\bm{\hat \lambda }}^{\rm{T}} - {\bm{\lambda }}^{\rm{T}}} \right){\bm{\delta }}/{k_{{\rm{d,}}i}} -\varepsilon_{e,i}/{k_{{\rm{d,}}i}}
	\label{eq3_10}
\end{equation}

Define a Lyapunov function as \eqref{eq3_11},
\begin{equation}
	V = \frac{1}{2}{k_{{\rm{p,}}i}}{k_{{\rm{d,}}i}}e_{^i}^2 + \frac{1}{2}\left( {{\bm{\hat \lambda }}^{\rm{T}} - {\bm{\lambda }}^{\rm{T}}} \right){{\bf{H}}}\left( {{{{\bm{\hat \lambda }}}} - {{\bm{\lambda }}}} \right)
	\label{eq3_11}
\end{equation}
where ${\bf{H}}$ is a positive definite diagonal gain matrix. Equation \eqref{eq3_10} can be subscribed into the derivative of the Lyapunov function \eqref{eq3_11}, and the result has a form as \eqref{eq3_12}.
\begin{equation}
	\begin{array}{l}
		\dot V = {k_{{\rm{p,}}i}}{k_{{\rm{d,}}i}}{e_i}{{\dot e}_i} + \left( {{{{\bm{\hat \lambda }}}^{\rm{T}}} - {{\bm{\lambda }}^{\rm{T}}}} \right){{\bf{H}}}{{{\bm{\dot {\hat \lambda} }}}}\\
		=   - {k^2_{{\rm{p,}}i}}e_{^i}^2 + \left( {{{{\bm{\hat \lambda }}}^{\rm{T}}} - {{\bm{\lambda }}^{\rm{T}}}} \right)\left( {{\bm{\delta }}{k_{{\rm{p,}}i}}{e_i} + {{\bf{H}}}{\bm{\dot {\hat \lambda} }}} \right) -{k_{{\rm{p,}}i}}{{e}_i}\varepsilon_{e,i}
	\end{array}
	\label{eq3_12}
\end{equation}

In the last line of \eqref{eq3_12}, The middle term can be eliminated by updating the parameter ${{\bm{{\hat \lambda} }}}$ through \eqref{eq3_13}. 
\begin{equation}
	{{\bm{\dot {\hat \lambda} }}}  =   -{k_{{\rm{p}},i}}{\bf{H}}^{-1}{\bm{\delta }}{e_i}
	\label{eq3_13}
\end{equation}
After \eqref{eq3_12} is simplified by \eqref{eq3_13}, the derivative of the Lyapunov function can be rewritten as \eqref{eq3_12b}.
\begin{equation}
	\begin{array}{l}
		\dot V =  - {k^2_{{\rm{p,}}i}}e_{^i}^2 -{k_{{\rm{p,}}i}}{{e}_i}\varepsilon_{e,i}
	\end{array}
	\label{eq3_12b}
\end{equation}
To stabilize the system, the relationship $- {k_{{\rm{p,}}i}}e_{^i}^2 < {{e}_i}\varepsilon_{e,i}$ must be guaranteed. Consider that $\varepsilon_{e,i}$ is bounded by a positive value $\mu$, the stabilization can be sufficiently ensured by satisfying $- {k_{{\rm{p,}}i}}e_{^i}^2 < -{\mu}|{{e}_i}|$, which is equivalent to $|e_{^i}| > {\mu}/{k_{{\rm{p,}}i}}$. That means the error can be converged to a neighbor region of zero whose length is ${\mu}/{k_{{\rm{p,}}i}}$, and this value can be close to zero with a sufficiently large ${k_{{\rm{p,}}i}}$.

Thereafter, the estimator can be updated by integrating \eqref{eq3_13}, and the estimator \eqref{eq3_13b} can be derived from \eqref{eq3_8b} and \eqref{eq3_13}.
\begin{equation}
	{{\tilde q}_{{\rm{d}},i}} = -\int {{{\bm{\delta }}^{\rm{T}}}{e_i}{\rm{d}}t \cdot{\bf{H}}^{-1}} {\bm{\delta }}
	\label{eq3_13b}
\end{equation}
Note that \eqref{eq3_13b} do not need the knowledge of the parameters in the PD joint position controller, so the PD controller and the estimator can be tuned separately.

\subsection{The Objective Tracking Controller}
\label{OTC}
The objective tracking controller feeds desired joint position ${{\mathbf{\bar q}}_{\rm{d}}}$ to the joint controller, and the desired position and orientation in task space is the controller input. MPC is formulated in the task space to address the joint constraints, and the optimization problem is formulated as \eqref{eq3_14a},
\begin{equation}
	\mathop {\min }\limits_{\bf{u}} \sum\limits_{k = 0}^{{N_{\rm{p}}}} {\left\{ {\left\| {{\bf{\tilde X}}[k] - {{{\bf{\tilde X}}}_{\rm{d}}}[k]} \right\|_{\bf{Q}}^2 + \left\| {{\bf{u}}[k]} \right\|_{\bf{R}}^2} \right\}} 
	\label{eq3_14a}
\end{equation}
%\begin{equation}
%	\mathop {\min }\limits_{{\mathbf{\dot q}}} \left\{ {{{\left\| {{\mathbf{J\dot q}} - {\mathbf{\dot x}}} \right\|}^2} + \left\| {{\mathbf{\dot q}}} \right\|_{\mathbf{W}}^2} \right\}
%	\label{eq3_14a}
%\end{equation}
%where ${\bm{\varepsilon }} = {[{\bm{\varepsilon }}_{\text{p}}^{\text{T}},{\bm{\varepsilon }}_{\text{o}}^{\text{T}}]^{\text{T}}}$ denotes the error in the task space, ${\bm{\varepsilon }}_{\text{p}},{\bm{\varepsilon }}_{\text{o}} \in {\mathbb{R}_3}$ denote the position and orientation errors. Let $\bm{p} \in \mathbb{R}_3$ denotes the end-effector position and a set of ZYX Euler angles $\bm{o} \in \mathbb{R}_3$ denotes the orientation. The position error is defined as the deviation between the actual position and desired position, ${\bm{\varepsilon }}_{\text{p}} =  \bm{p}-\bm{p}_{\rm{d}}$. The orientation error also has an analogous form as ${\bm{\varepsilon }}_{\text{o}} =  \bm{o}-\bm{o}_{\rm{d}}$. Equation \eqref{eq3_14a} minimizes the norm of the error in the task space and also minimizes the norm of the increment of the control command $\mathbf{u}$. ${\mathbf{Q}}$, ${\mathbf{R}} \in {\mathbb{R}_{6 \times 6}}$ are two weight matrices. $N_{\text{p}}$ is the length of the prediction horizon of the controller. The end-effector in the task space satisfies the dynamics \eqref{eq3_14b},
where ${\bf{\tilde X}} = {[{{\bf{\tilde p}}^{\rm{T}}},{{\bf{\tilde o}}^{\rm{T}}}]^{\rm{T}}}$ denotes the state deviation from the current pose, ${\mathbf{Q}}$, ${\mathbf{R}} \in {\mathbb{R}_{6 \times 6}}$ are two weight matrices. $N_{\text{p}}$ is the length of the prediction horizon of the controller. ${\bf{u}} = {[{{\bf{v}}^{\rm{T}}}\;{{\bf{w}}^{\rm{T}}}]^{\rm{T}}}$ is solved in real-time, and ${\bf{v}}$ and ${\bf{w}}$ is the translation and angular velocity in the task space.

Let $\bf{p} \in \mathbb{R}_3$ denotes the current end-effector position and a quaternion $\bm{\xi } \in \mathbb{R}_4$ denotes the orientation. The desired position and orientation deviation ${{{\bf{\tilde X}}}_{\rm{d}}} = {[{\bf{\tilde p}}_{\rm{d}}^{\rm{T}},{\bf{\tilde o}}_{\rm{d}}^{\rm{T}}]^{\rm{T}}}$ is defined as \eqref{eq3_14s1} and \eqref{eq3_14s2},
\begin{subequations}
	\begin{equation}
		{{{\bf{\tilde p}}}_{\rm{d}}}[k] = {{\bf{p}}_{\rm{d}}}[k] - {\bf{p}}
		\label{eq3_14s1}
	\end{equation}
	\begin{equation}
		[\xi_{\rm{d}} [k]\;{{{\bm{\tilde o}}}_{\rm{d}}}[k]] = {{\bm{\xi }}_{\rm{d}}}[k]{{\bm{\xi }}^{ - 1}}
		\label{eq3_14s2}
	\end{equation}
\end{subequations}
where $\xi_{\rm{d}}$ is the real part of the quaternion, ${{{\bf{ p}}}_{\rm{d}}}$ and ${{{\bm{\xi}}}_{\rm{d}}}$ is the desired position and orientation. Equation \eqref{eq3_14a} minimizes the norm of the error in the task space and also minimizes the norm of the control command $\mathbf{u}$.  

In each control period, the manipulator satisfies the dynamics \eqref{eq3_14b} in the task space,
\begin{equation}
	{\bf{\tilde X}}[k + 1] = {\bf{\tilde X}}[k] + \Delta t{\bf{u}}[k]
	\label{eq3_14b}
\end{equation}
where $\Delta t$ is the control period, $\mathbf{u} \in \mathbb{R}_6$ is the velocity command in the task space, which is constrained by the joint state limits, as shown in \eqref{eq3_14c},
\begin{equation}
	{g_{\text{l}}}({\bf{q}}) \le {{{\bf{J}}^\# }{\bf{u}}[k]} \le {g_{\text{u}}}({\bf{q}})
	\label{eq3_14c}
\end{equation}
where ${{\bf{J}}^\# }$ is the pseudo-inverse of the geometric Jacobian $\bf{J}$, $\bf{q}$ is the current joint position. ${g_{\text{l}}}$ and ${g_{\text{u}}}$ are the boundaries that dynamically change according to the distance to the joint position limitations. Note that the Jacobian varies with the joint position in the prediction horizon, which is difficult to be linearized. Therefore, the constraint \eqref{eq3_14c} is relaxed by assuming the Jacobian to be constant in the prediction horizon. This assumption is also applied in \cite{wahrburg2016mpc, woolfrey2021predictive}. Although the optimality cannot be guaranteed by this method, the solution for application is still suboptimal and feasible.

The joint states of our manipulator are limited by position, velocity, acceleration, and jerk, which can be expressed by \eqref{eq3_15a}-\eqref{eq3_15d},
\begin{subequations}
	\begin{equation}
		{{\bf{q}}_{\rm{l}}} \le {\bf{q}} \le {{\bf{q}}_{\rm{u}}}
		\label{eq3_15a}
	\end{equation}
	\begin{equation}
		{{{\bf{\dot q}}}_{\rm{l}}} \le {\bf{\dot q}} \le {{{\bf{\dot q}}}_{\rm{u}}}
		\label{eq3_15b}
	\end{equation}
	\begin{equation}
		{{{\bf{\ddot q}}}_{\rm{l}}} \le {\bf{\ddot q}} \le {{{\bf{\ddot q}}}_{\rm{u}}}
		\label{eq3_15c}
	\end{equation}
	\begin{equation}
		{{{\bf{\dddot q}}}_{\text{l}}} \le {\bf{\dddot q}} \le {{{\bf{\dddot q}}}_{\text{u}}}
		\label{eq3_15d}
	\end{equation}
\end{subequations}
where ${{\bf{q}}_{\rm{u}}}$ and ${{\bf{q}}_{\rm{l}}}$ are the upper and lower limitations of the joint position. 

The joint state constraints \eqref{eq3_15a}-\eqref{eq3_15d} can be transformed to joint velocity constraint to satisfy the condition in \eqref{eq3_14c}. Similar handling of velocity and acceleration constraints is utilized in \cite{flacco2015control}, but jerk is not considered. The velocity constraint is acquired in four different conditions. The first condition is that the joint position in the neighbor of constraints, and \eqref{eq3_16} is satisfied.
\begin{equation}
	{{\bf{q}}_{\text{l}}} - {\bf{q}} \le \Delta t{\bf{\dot q}} \le {{\bf{q}}_{\text{u}}} - {\bf{q}}
	\label{eq3_16}
\end{equation}

The second condition is that velocity reaches the limitations, and the velocity constraint is the same as \eqref{eq3_15b}. The third condition is that the manipulator runs at maximum acceleration. In this condition, the velocity should satisfy\eqref{eq3_17a} and \eqref{eq3_17b}.
\begin{subequations}
	\begin{equation}
		{{{{\bf{\ddot q}}}_{\rm{u}}}{t^2} = 2\left( {{{\bf{q}}_{\rm{u}}} - {\bf{q}}} \right)}
		\label{eq3_17a}
	\end{equation}
	\begin{equation}
		{\bf{\dot q}} \le {{{\bf{\ddot q}}}_{\rm{u}}}t
		\label{eq3_17b}
	\end{equation}
\end{subequations}

The last condition is that the jerk of the manipulator reaches the limitation, and the velocity should satisfy \eqref{eq3_18a} and \eqref{eq3_18b}.
\begin{subequations}
	\begin{equation}
		{{{\bf{\dddot q}}}_{\text{u}}}{t^3} = 6\left( {{{\bf{q}}_{\rm{u}}} - {\bf{q}}} \right)
		\label{eq3_18a}
	\end{equation}
	\begin{equation}
		{\mathbf{\dot q}} \le \frac{1}{2}{{{\bf{\dddot q}}}_{\text{u}}}{t^2}
		\label{eq3_18b}
	\end{equation}
\end{subequations}
According to these conditions, the maximum velocity is the minimum value of the velocities acquired by equation \eqref{eq3_15b} and \eqref{eq3_16} to \eqref{eq3_18b}. The maximum velocity can be expressed by \eqref{eq3_19},
\begin{equation}
	{g_{\text{u}}}({\mathbf{q}}) = \min \left\{ {\frac{{\Delta {{\mathbf{q}}_u}}}{\Delta t},{{{\mathbf{\dot q}}}_{\text{u}}},\sqrt {2{{{\mathbf{\ddot q}}}_{\text{u}}}\Delta {{\mathbf{q}}_u}} ,\sqrt[3]{{\frac{9}{2}{{{\mathbf{\dddot q}}}_{\text{u}}}\Delta {\mathbf{q}}_{\text{u}}^2}}} \right\}
	\label{eq3_19}
\end{equation}
where $\Delta {{\mathbf{q}}_{\text{u}}} = {{\mathbf{q}}_{\text{u}}} - {\mathbf{q}}$. Let $\Delta {{\mathbf{q}}_{\text{l}}} = {{\mathbf{q}}_{\text{l}}} - {\mathbf{q}}$, The minimum velocity can be acquired by the similar method, which is expressed by \eqref{eq3_20}.
\begin{equation}
	{g_{\text{l}}}({\mathbf{q}}) = \max \left\{ {\frac{{\Delta {{\mathbf{q}}_{\text{l}}}}}{\Delta t},{{{\mathbf{\dot q}}}_{\text{l}}}, - \sqrt {2{{{\mathbf{\ddot q}}}_{\text{l}}}\Delta {{\mathbf{q}}_{\text{l}}}} ,\sqrt[3]{{\frac{9}{2}{{{\mathbf{\dddot q}}}_{\text{l}}}\Delta {\mathbf{q}}_{\text{l}}^2}}} \right\}
	\label{eq3_20}
\end{equation}
According to \eqref{eq3_19} and \eqref{eq3_20}, the velocity is bounded in a closed region, and the maximum velocity decreases smoothly when the joint position is getting closer to the boundary. 

The optimization problem \eqref{eq3_14a}-\eqref{eq3_14c} can be formulated to a QP problem, which can be solved in real-time by the toolbox like OSQP \cite{osqp}. The optimal control sequence is solved in every control period, and only the first component ${\bf{u}}^*_0$ will be applied as the control command. 

The desired joint velocity is acquired by the inverse kinematics in \eqref{eq3_21}.
\begin{equation}
	{\mathbf{\dot q}} = {{\mathbf{J}}^{\text{T}}}{(\rho {\mathbf{I}} + {\mathbf{J}}{{\mathbf{J}}^{\text{T}}})^{{{ - 1}}}}{{{\bf{u}}^*_0}}
	\label{eq3_21}
\end{equation}
Considering that passive base motion may lead to unexpected joint configuration, a sufficiently small parameter $\rho$ can avoid the ill condition. The desired joint position command ${{\mathbf{\bar q}}_{\rm{d}}}$ can be acquired by integrating the joint velocity in \eqref{eq3_21}. In the traditional method, to track the objective in the world frame, the velocity of end-effector in task space can be defined as ${\mathbf{\dot x}} = {{{\bf{\dot x}}}_{\rm{d}}}+{\mathbf{K}}\left( {{{\mathbf{x}}_{\text{d}}} - {\mathbf{x}}} \right)$ \cite{flacco2015control}. ${\mathbf{K}}$ is a positive definite diagonal gain matrix that makes the tracking error converges to zero exponentially.

\subsection{UAV Position Feedback}
\label{UPF}
\begin{figure}[t]
	\centering
	\setlength{\abovecaptionskip}{0pt}
	\setlength{\belowcaptionskip}{0pt}
	\includegraphics[width=1\linewidth]{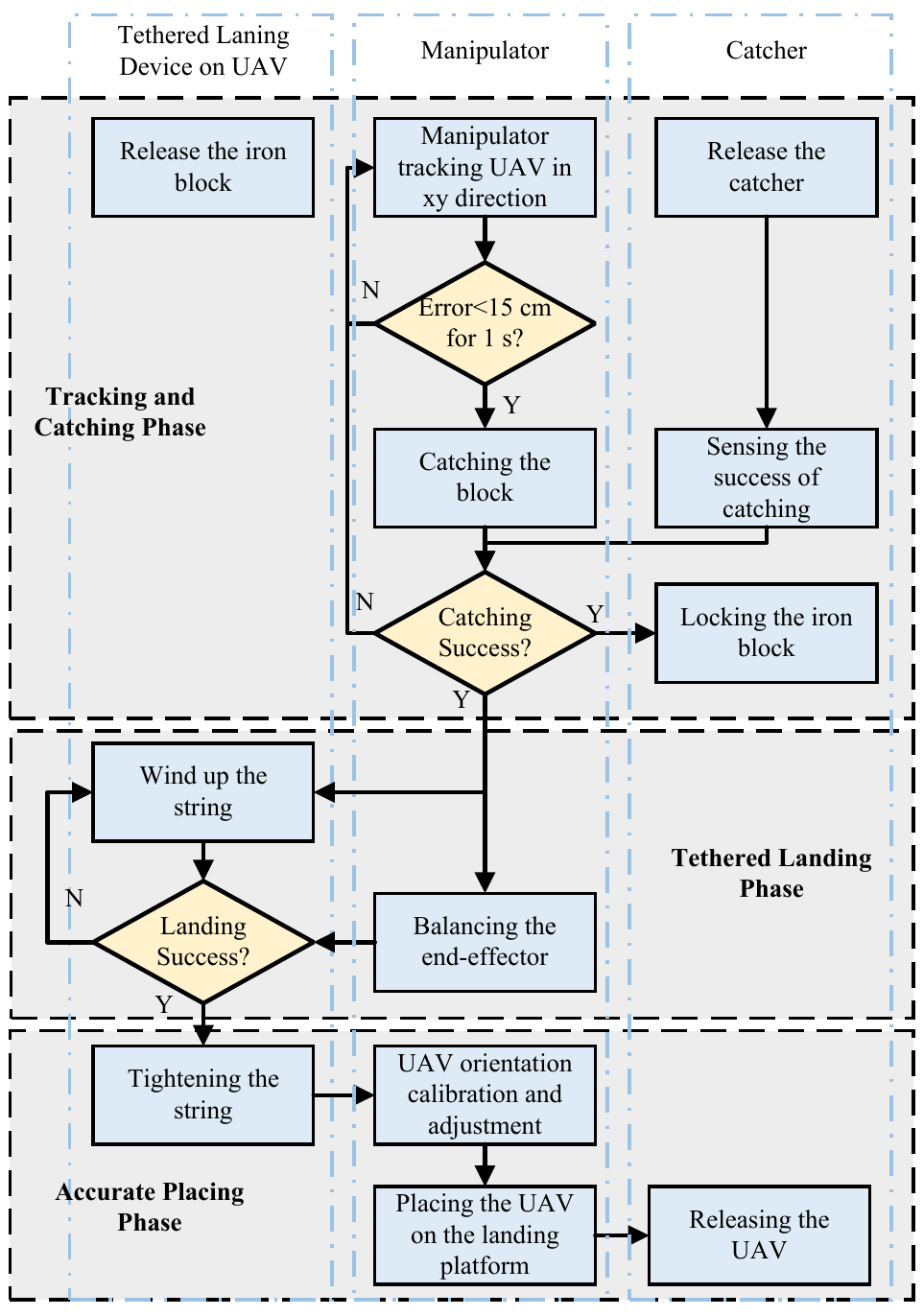}
	\caption{Control strategy of the landing assistance system.}
	\label{ctl_stra}
\end{figure}
The position of the UAV can be captured by the infrared camera system, and marks are attached to the UAV to reflect the infrared rays. The IR cameras emit and receive 850 nm infrared, but they can be interfered with by ambient light. Reducing the camera exposure can mitigate this problem because the infrared rays reflected by the marks have higher intensity, but noises still exist at noon on a sunny day. The Kalman filter is applied to solve this problem.

Define the state of the UAV in the camera frame as ${{\bf{x}}_a}$.
\begin{equation}
	{{\bf{x}}_a} = [{x_a}\;{y_a}\;{z_a}\;{\dot x_a}\;{\dot y_a}\;{\dot z_a}]
	\label{eq3_a_23}
\end{equation}
In \eqref{eq3_a_23}, $x_a$, $y_a$, and $z_a$ denote the position of the UAV in the camera frame. Define the control period as $\Delta t$, the discrete-time dynamic model of the UAV can be described by \eqref{eq3_a_24a} and \eqref{eq3_a_24b},
\begin{subequations}
	\begin{equation}
		{{\bf{x}}_a}[k + 1] = {\bf{D}}{{\bf{x}}_a}[k] + {\bf{w}}[k]
		\label{eq3_a_24a}
	\end{equation}
	\begin{equation}
		{{\bf{y}}_a}[k] = {\bf{F}}{{\bf{x}}_a}[k] + {\bf{v}}[k]
		\label{eq3_a_24b}
	\end{equation}
\end{subequations}
where $\bf{D}$ and $\bf{F}$ are state transition matrix and measurement matrix, respectively. ${\bf{y}}_a$ is the measured position. ${\mathbf{w}} \sim N\left( {0,{\mathbf{Q_{\rm{S}}}}} \right)$ represents zero-mean system Gaussian noise whose covariance matrix is ${\mathbf{Q_{\rm{S}}}}$, and ${\mathbf{v}} \sim N\left( {0,{\mathbf{Q_{\rm{M}}}}} \right)$ is the zero-mean measurement Gaussian noise noise with covariance matrix ${\mathbf{Q_{\rm{M}}}}$. $\bf{D}$ and $\bf{F}$ are given by \eqref{eq3_a_25}.
\begin{equation}
	{\mathbf{D}} = \left[ {\begin{array}{*{20}{c}}
			{{{\mathbf{I}}_{3 \times 3}}}&{\Delta t{{\mathbf{I}}_{3 \times 3}}} \\ 
			{{{\mathbf{0}}_{3 \times 3}}}&{{{\mathbf{I}}_{3 \times 3}}} 
	\end{array}} \right],\;{\mathbf{F}} = \left[ {\begin{array}{*{20}{c}}
			{{{\mathbf{I}}_{3 \times 3}}}&{{{\mathbf{0}}_{3 \times 3}}} 
	\end{array}} \right]
	\label{eq3_a_25}
\end{equation}

${{{\mathbf{I}}_{n \times n}}}$ denotes $n$-dimension identity matrix. ${\mathbf{Q_{\rm{S}}}}$ and ${\mathbf{Q_{\rm{M}}}}$ are related to the control period $\Delta t$, which are given by \eqref{eq3_a_26}.
\begin{equation}
	{{\mathbf{Q_{\rm{S}}}}} = \left[ {\begin{array}{*{20}{c}}
			{\Delta {t^3}{{\mathbf{I}}_{3 \times 3}}}&{\Delta {t^2}{{\mathbf{I}}_{3 \times 3}}} \\ 
			{\Delta {t^2}{{\mathbf{I}}_{3 \times 3}}}&{\Delta t{{\mathbf{I}}_{3 \times 3}}} 
	\end{array}} \right],\;{{\mathbf{Q_{\rm{M}}}}} = \Delta t{{\mathbf{I}}_{6 \times 6}}
	\label{eq3_a_26}
\end{equation}
Thereafter, the robust Kalman filter can be applied by following the same steps as \cite{kai2009robust}. 

\subsection{System Control Strategy}
\label{SCS}

This section introduces the system control strategy for landing assistance when the UAV is in the visual field of the camera system. The landing assistance process consists of three phases, the tracking and catching phase, the tethered landing phase, and the accurate placing phase. The process is illustrated in Fig. \ref{ctl_stra}.

%In the tracking and catching phase, the manipulator will track the UAV in an available zone ${\Omega _a}$ in the base frame of the manipulator, which is a subset of the dexterous space ${\Omega _d}$ of the manipulator. The x 
%A surface is designed to ensure the manipulator can work in the dexterous space
%The manipulator end-effector will slide on a surface in the available zone while tracking the UAV. The sliding surface is preset to ensure the manipulator work in the dexterous space. The manipulator will catch the block released by the UAV as long as the tracking error can be kept within 0.03 m for two seconds.

In the tracking and catching phase, the manipulator will track the UAV in the $x$ and $y$ direction in the manipulator base frame. A tracking space is defined in \eqref{eq3_b_26} to ensure the manipulator work in the dexterous space,
\begin{equation}
	{\Omega _{\text{t}}} = \left\{ {x,y,z:\left\| {[x,y]} \right\| \leqslant r,{S_L}\left( {x,y} \right) \leqslant z \leqslant {S_U}\left( {x,y} \right)} \right\}
	\label{eq3_b_26}
\end{equation}
where $r$ is the maximum catching range, ${S_L}$ and ${S_U}$ are the lower and upper boundary functions in the $z$ direction, which is preset according to the manipulator structure. The UAV will hover above the manipulator and release the docking block to the catching zone. Because the released string has a fixed length, the controller can estimate the block position by the UAV position provided by the camera system. 

The manipulator will try to catch the docking block if a tracking error less than 15 cm can be kept for more than 1 seconds, and the error is defined as the norm of the error in the $x$ and $y$ direction. During the tracking process, the catcher will release the PVC strings. The catcher has a sufficiently large face equipped with magnets, which improves the success rate of the catching process. The catcher can detect the success of the catching operation, as introduced in Section \ref{catcher}. If the catch fails to catch the block, the manipulator will repeat the tracking and catching process until the catcher attracts the block successfully. After that, the block will be locked on the catcher in a minute by the PVC strings, and the tethered landing operation will be executed.

In the tethered landing phase, the manipulator will balance the end-effector for UAV landing. The tethered system winds up the tether to dock the UAV on the locked docking block. The tethered landing system can detect the success of docking. The system will keep winding up the string until the signal of success docking is detected. After that, the UAV will be tightened, and no relative movement between the catcher and the UAV will be generated.

In the accurate landing phase, the manipulator will track a preset trajectory to place the UAV on the landing platform. However, the UAV can dock on the manipulator with a different orientation, so the orientation of the UAV must be calibrated before executing the trajectory for accurate placing. In the calibration process, the manipulator will move the UAV to the front of the camera system. After that, the manipulator will adjust the UAV pose to the same as the initial pose of the preset trajectory. Thereafter, the UAV will be placed on the landing platform. The landing platform can fix the UAV on the USV or release the UAV for new tasks.

\section{Simulations and Experiments}
\label{exp}
Simulations and experiments are conducted and introduced in this section. We first verify the feasibility of the manipulator controller by simulations. After that, indoor experiments are conducted to test the performance of the controller. The functionality of the landing assistance system is verified by field experiments on the water surface. Parameters of the system and controller are summarized in Table \ref{table1}. 
\begin{table}[t]
	\centering
	\caption{System Parameters and Control Parameters}
	\label{table1}
	\begin{tabular}{lllll}
		\toprule[1.5pt]
		\multicolumn{5}{l}{\begin{tabular}[c]{@{}l@{}}$\bf{Manipulator \  Parameters:}$\\
				$\rm{Catching \ range} r: 700 \ mm, \ \rm{Working \ load:}3 \ kg$,\\
				${{{\bf{ q}}}_{\rm{u}}}=-{{{\bf{ q}}}_{\rm{l}}}=[175, 120, 175, 120, 175, 120, 360] \ \rm{degs}$, \\ 
				${{{\bf{\dot q}}}_{\rm{u}}}=-{{{\bf{\dot q}}}_{\rm{l}}}=[180, 180, 180, 180, 225, 225, 225] \ \rm{degs/s}$, \\
				${{{\bf{\ddot q}}}_{\rm{u}}}=-{{{\bf{\ddot q}}}_{\rm{l}}}=[900, 900, 900, 900, 1125, 1125, 1125] \ \rm{degs/s^2}$. \\
				${{{\bf{\dddot q}}}_{\rm{u}}}=-{{{\bf{\dddot q}}}_{\rm{l}}}=[18, 18, 18, 18, 22.5, 22.5, 22.5] \times 1000 \ \rm{degs/s^3}$ \\
				$\bf{Catcher \  Parameters:}$\\
				$\rm{Magnet \ size}:10 \times 5 \ mm$, $\rm{Magnets \ number}: 120$ \\
				Catcher \ locking \ time: \ 0.8 s
		\end{tabular}} \\ 
		\midrule
		\multicolumn{5}{l}{\begin{tabular}[c]{@{}l@{}}$\bf{Control \  \ Parameters:}$\\ 
				${{\bf{K}}_{\rm{p}}} = diag\left\{150,150,150,150,70,70,70\right\}$,\\ 
				${{\bf{K}}_{\rm{d}}} = diag\left\{50,50,50,50,20,20,20\right\}$,\ ${{\bf{H}}} = diag\left\{1,0.01,10\right\}$\\
				${{\bf{K}}} = diag\left\{60,60,60,40,40,40\right\}$,\ ${\rho} = 0.01 $\\
				${\bf{Q}}=100{\bf{I}},\ {\bf{R}}=0.1{\bf{I}}, \ \Delta t=0.01 \ s$ 
		\end{tabular}}                                         \\ \bottomrule[1.5pt]
	\end{tabular}
\end{table}

\subsection{Simulation Results of Manipulator Controller}
\begin{figure}[t]
	\centering
	\setlength{\abovecaptionskip}{0pt}
	\setlength{\belowcaptionskip}{0pt}
	\includegraphics[width=3.3in]{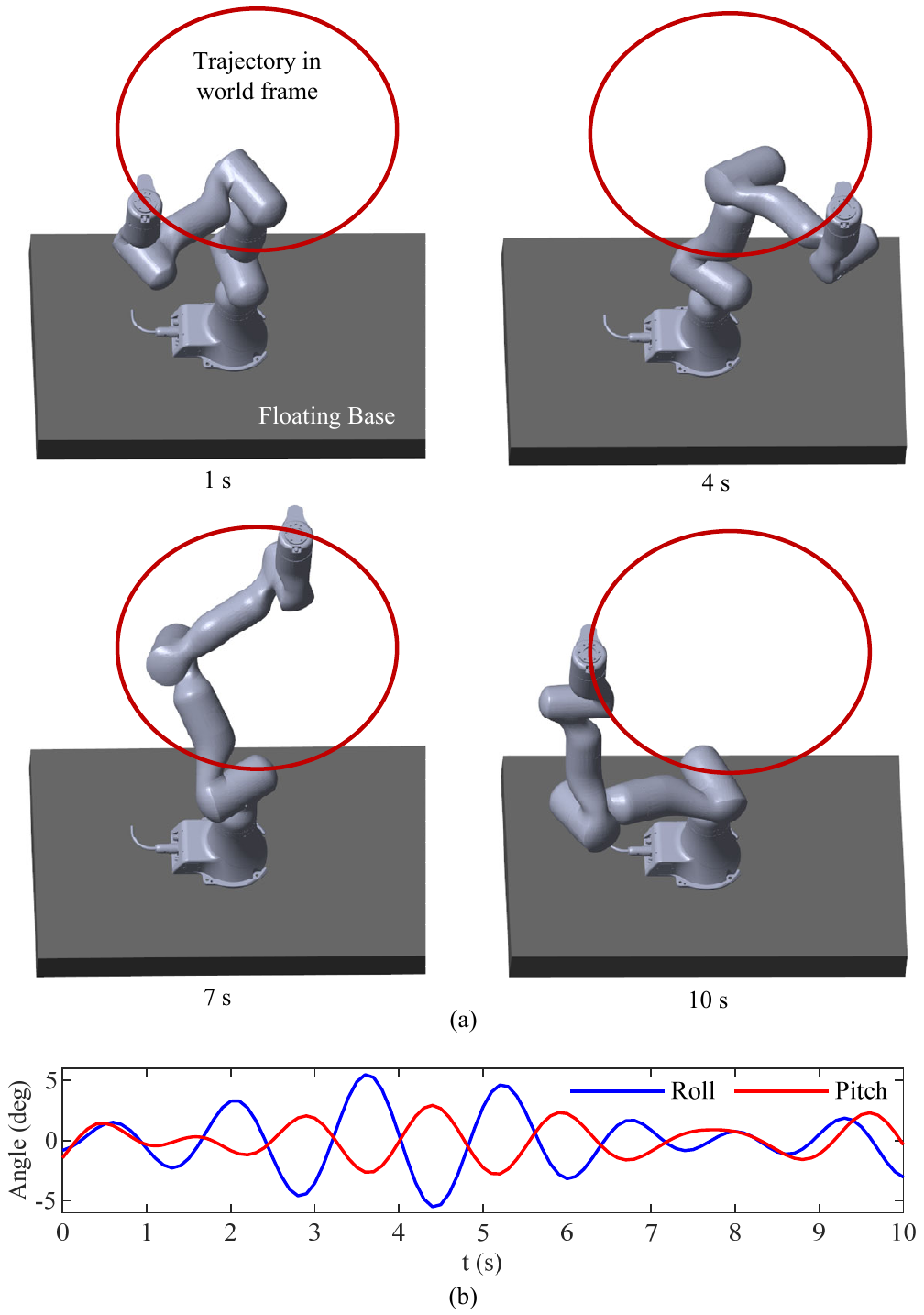}
	\caption{Simulation environment and setup. (a) Manipulator physical model in Simscape and the tracking result of a circular trajectory in the world frame. (b) The simulated base motion.}
	\label{simset}
\end{figure}

\begin{figure}[t]
	\centering
	\setlength{\abovecaptionskip}{0pt}
	\setlength{\belowcaptionskip}{0pt}
	\includegraphics[width=3.3in]{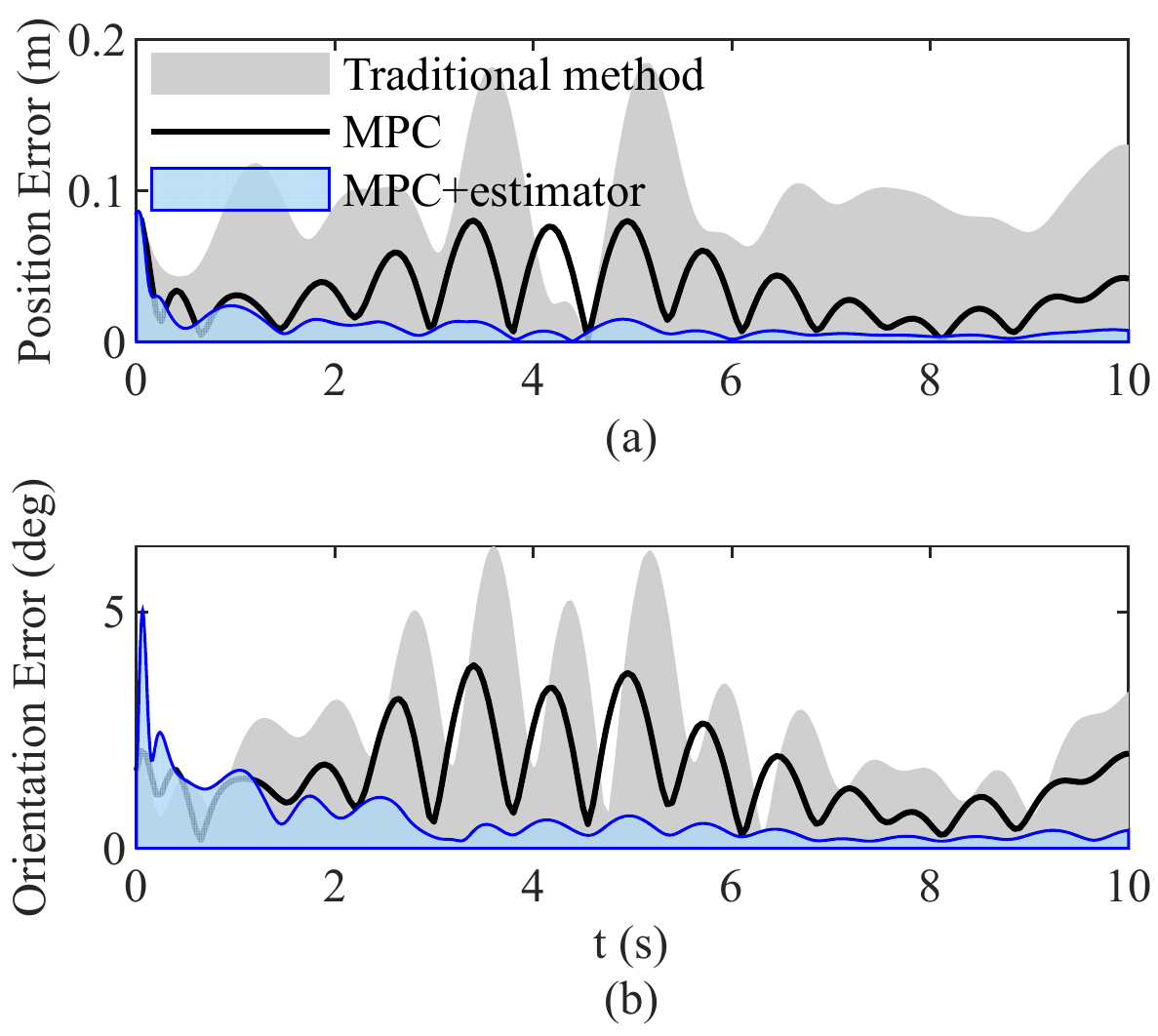}
	\caption{Norm of Errors in Simulation. (a) Norm of position error (b) Norm of orientation error.}
	\label{sim1_2}
\end{figure}
Simulations are conducted to test the feasibility of the manipulator controller. The simulator is designed based on Matlab/Simscape and Robotics System Toolbox (RST). The model in Simscape is shown in Fig. \ref{simset}(a), and the manipulator is mounted on a floating base whose motion is shown in Fig. \ref{simset}(b), which is collected from the real UAV.

The manipulator end-effector is made to track a circular trajectory in the world frame and balance the base motion at the same time. The radius of the desired trajectory is 0.35 m, and the whole trajectory is in the dexterous space of the floating manipulator. The controller is designed in Matlab/Simulink. The inverse kinematics is calculated in real-time based on the Jacobian matrix provided by RST. The joint position controller is a PD controller with gravity compensation. The Simscape model receives the joint torque command and provides feedback on joint position and velocity.

The results are shown in Fig. \ref{simset}(a). The tracking performance of the MPC controller with an estimator is compared with a traditional controller and an MPC controller. The PD joint position controller is applied in all the three cases, in which ${\bf{K}}_{\rm{p}}$ and ${\bf{K}}_{\rm{d}}$ are $diag\left\{150,150,150,150,70,70,70\right\}$ and $diag\left\{50,50,50,50,20,20,20\right\}$. The parameter of estimator is tuned as $diag\left\{1,0.01,10\right\}$. Fig. \ref{sim1_2}(a) shows the norm of three-dimension position errors and the orientation errors are shown in Fig. \ref{sim1_2}(b). The average errors of position and orientation are 0.092 m and 2.537 degrees for the traditional controller, and meanwhile, the maximum errors reach 0.185 and 6.387 degrees. The average errors of the MPC controller are 0.033 m and 1.563 degrees, and the maximum error is reduced to 0.086 m and 3.875 degrees. After the estimator is applied, the maximum error is 0.087 m and 5.038 degrees, and the average error is reduced to 0.024 m and 0.926 degrees, which improves the position and orientation tracking accuracy by 26.9$\%$ and 40.8$\%$. The error of the MPC with an estimator could be higher at the beginning of the tracking mission, but the average error is still lower than the controller without an estimator.

\subsection{Indoor Experiments of Manipulator Controller}
\begin{figure}[t]
	\centering
	\setlength{\abovecaptionskip}{0pt}
	\setlength{\belowcaptionskip}{0pt}
	\includegraphics[width=1\linewidth]{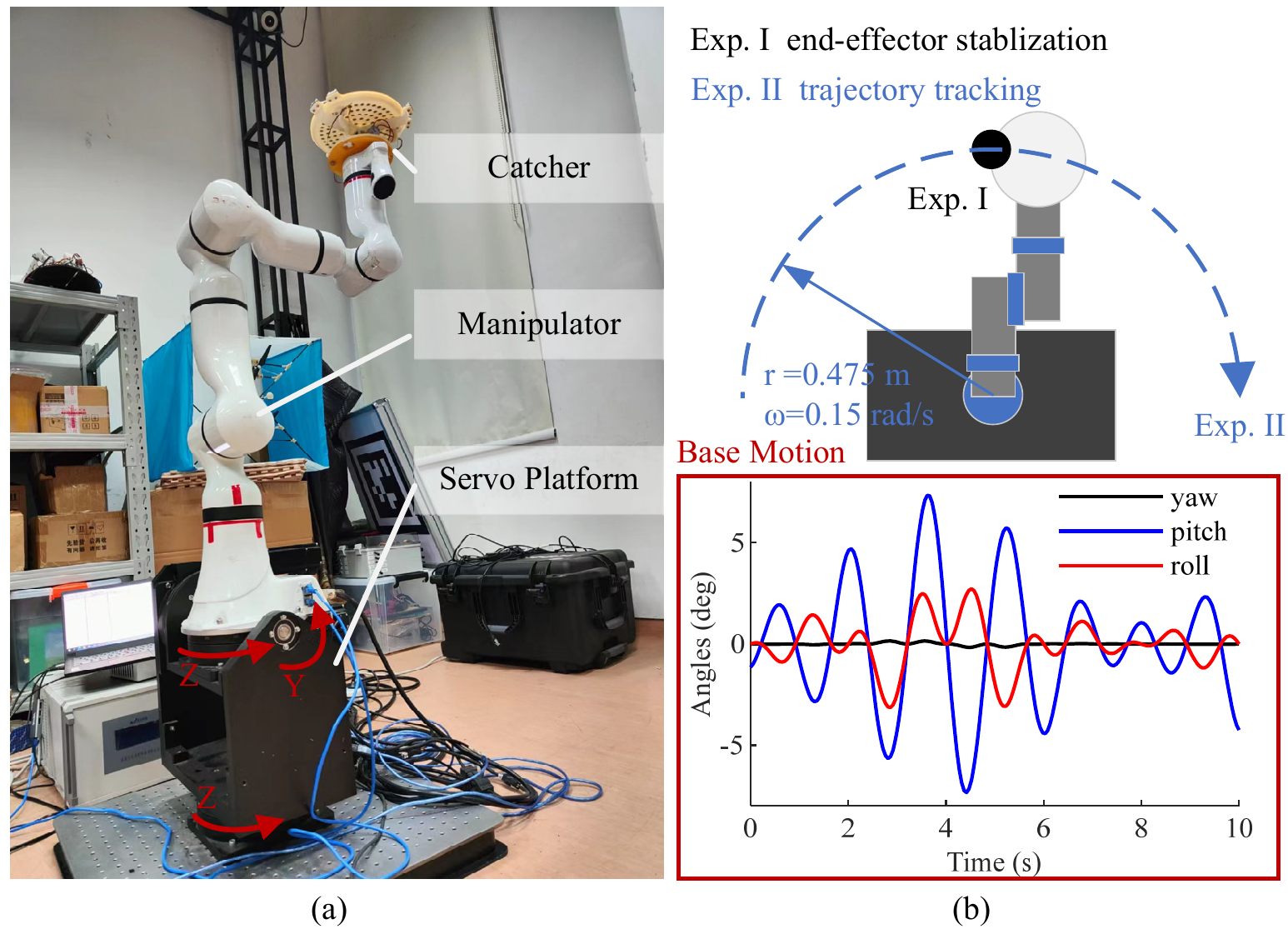}
	\caption{The indoor experimental setup. The experimental platform is shown in (a). (b) gives the illustration of indoor experiment design.}
	\label{exp_set}
\end{figure}
\begin{figure}[t]
	\centering
	\setlength{\abovecaptionskip}{0pt}
	\setlength{\belowcaptionskip}{0pt}
	\includegraphics[width=0.8\linewidth]{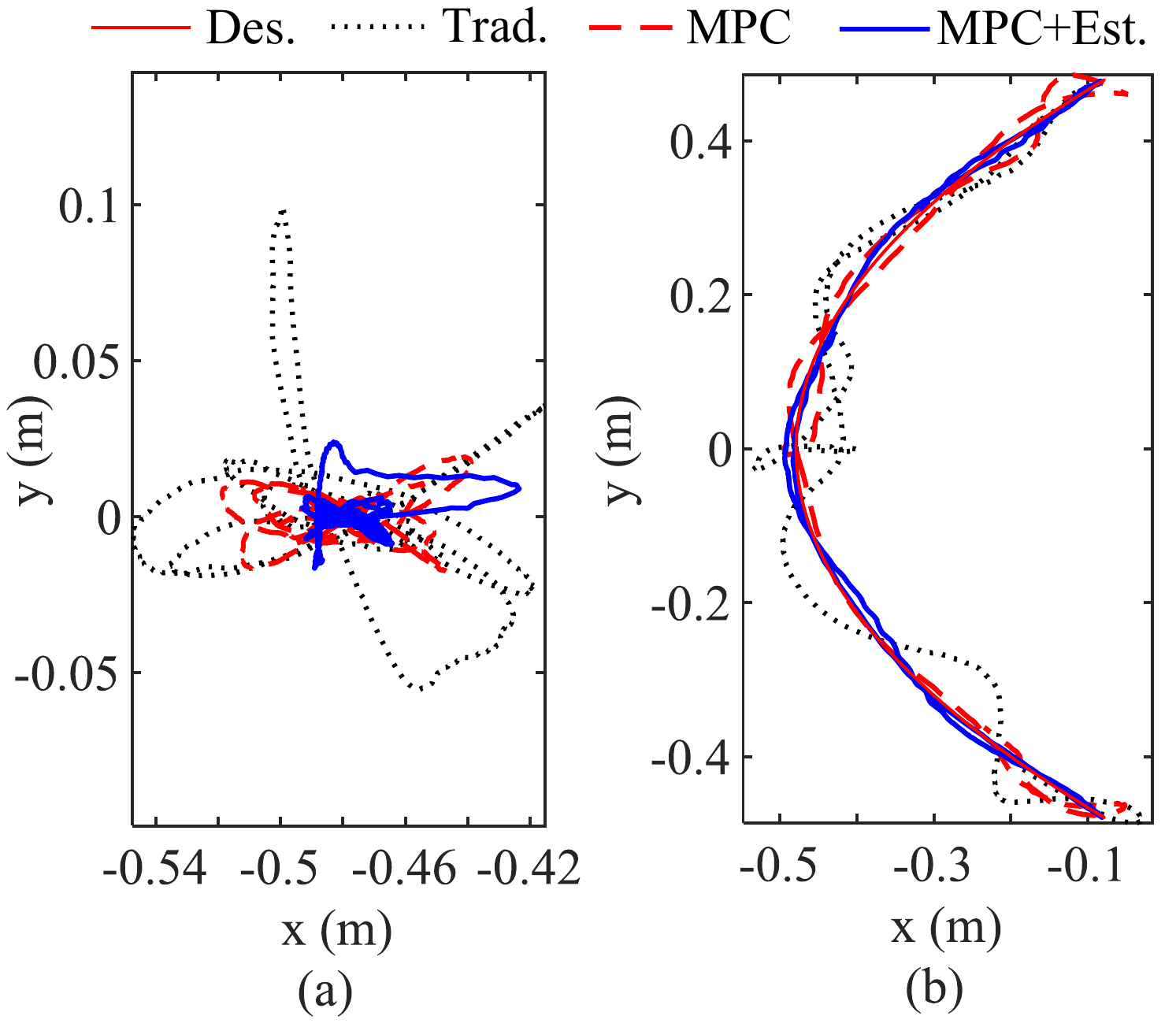}
	\caption{The comparison of end-effector trajectories in (a) the end-effector stabilization experiments and (b) the trajectory tracking experiments. The solid red line is the desired (Des.) trajectory,  and the dotted black, dash red, and solid blue line denotes the trajectory of the traditional (Trad.), MPC, and MPC with the estimator (Est.) method, respectively.}
	\label{exp_id_traj}
\end{figure}
The indoor experiments are conducted to verify the performance of the controller further. The experimental setup is shown in Fig. \ref{exp_set}(a). The manipulator is mounted on a rotary servo platform. The platform is programmed on Labview to simulate the motion of the USV. As shown in Fig. \ref{exp_set}(b), two groups of experiments are conducted to compare the performance of a traditional controller, an MPC controller with and without the estimator. The controller parameters are given in Table \ref{table1}. The simulated motion is described by the ZYX Euler angle, which is transformed to ZYZ Euler angle to fit the ZYZ structure of the servo platform. An IMU is mounted on the servo platform to provide the base motion feedback.  

\begin{figure}[t]
	\centering
	\setlength{\abovecaptionskip}{0pt}
	\setlength{\belowcaptionskip}{0pt}
	\includegraphics[width=1\linewidth]{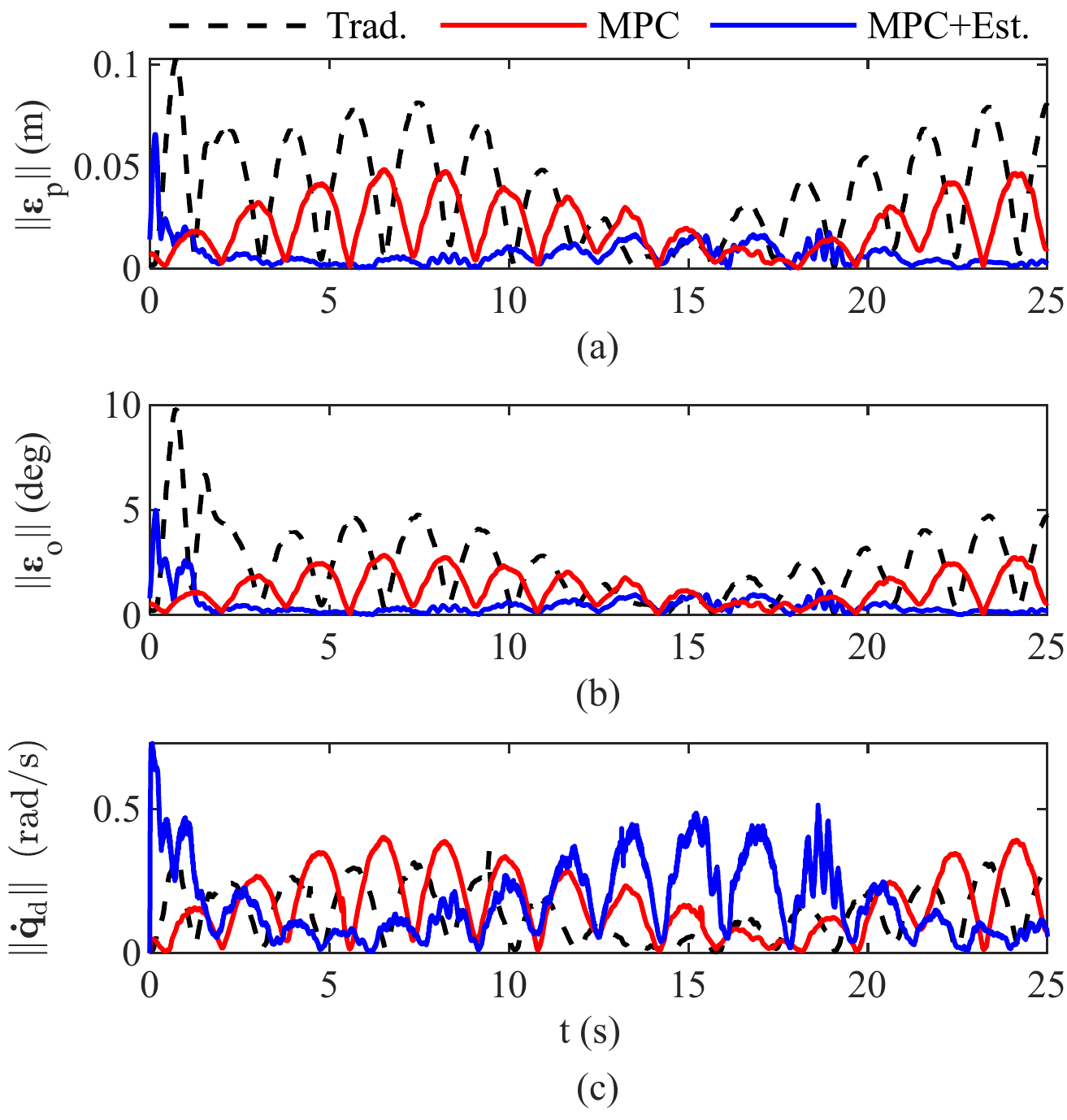}
	\caption{The indoor experimental results of end-effector stabilization. (a), (b), and (c) are the norm of the position errors, orientation errors, and task controller output, respectively.}
	\label{exp_id}
\end{figure}

In the first experiments, the manipulator is controlled to stabilize its end-effector at a point in the world frame. The trajectories in the XY plane are shown in Fig. \ref{exp_id_traj}(a), the result shows that the MPC can improve the accuracy of the end-effector stabilization, and after the estimator is applied, the control accuracy can be further improved. More details are shown in Fig. \ref{exp_id}(a). The average position error of the traditional controller is 0.037 m, and the maximum error is 0.102 m. The MPC controller without an estimator reduces the average error to 0.020 m, and the maximum error is reduced to 0.049 m. After the estimator is applied, the average error is reduced to 0.007 m, which improves 65$\%$, and the maximum error is 0.066 m. In Fig. \ref{exp_id}(b), the average orientation error of the traditional method is 2.356 degrees, and the average error of the MPC controller is 1.220 degrees. The average error is reduced to 0.459 degrees after the application of the estimator, which improves 62$\%$. The maximum error is reduced from 9.840 degrees to 2.852 degrees by the MPC controller, and the maximum error of the MPC with an estimator is 5.002 degrees. The control outputs of the three methods are shown in Fig. \ref{exp_id}(c). The maximum velocity outputs are close to each other, but the output changes more sensitively after the estimator is applied. The estimator can react to the variation of unknown dynamics, and additional effort is made for stabilization.

\begin{figure}[t]
	\centering
	\setlength{\abovecaptionskip}{0pt}
	\setlength{\belowcaptionskip}{0pt}
	\includegraphics[width=1\linewidth]{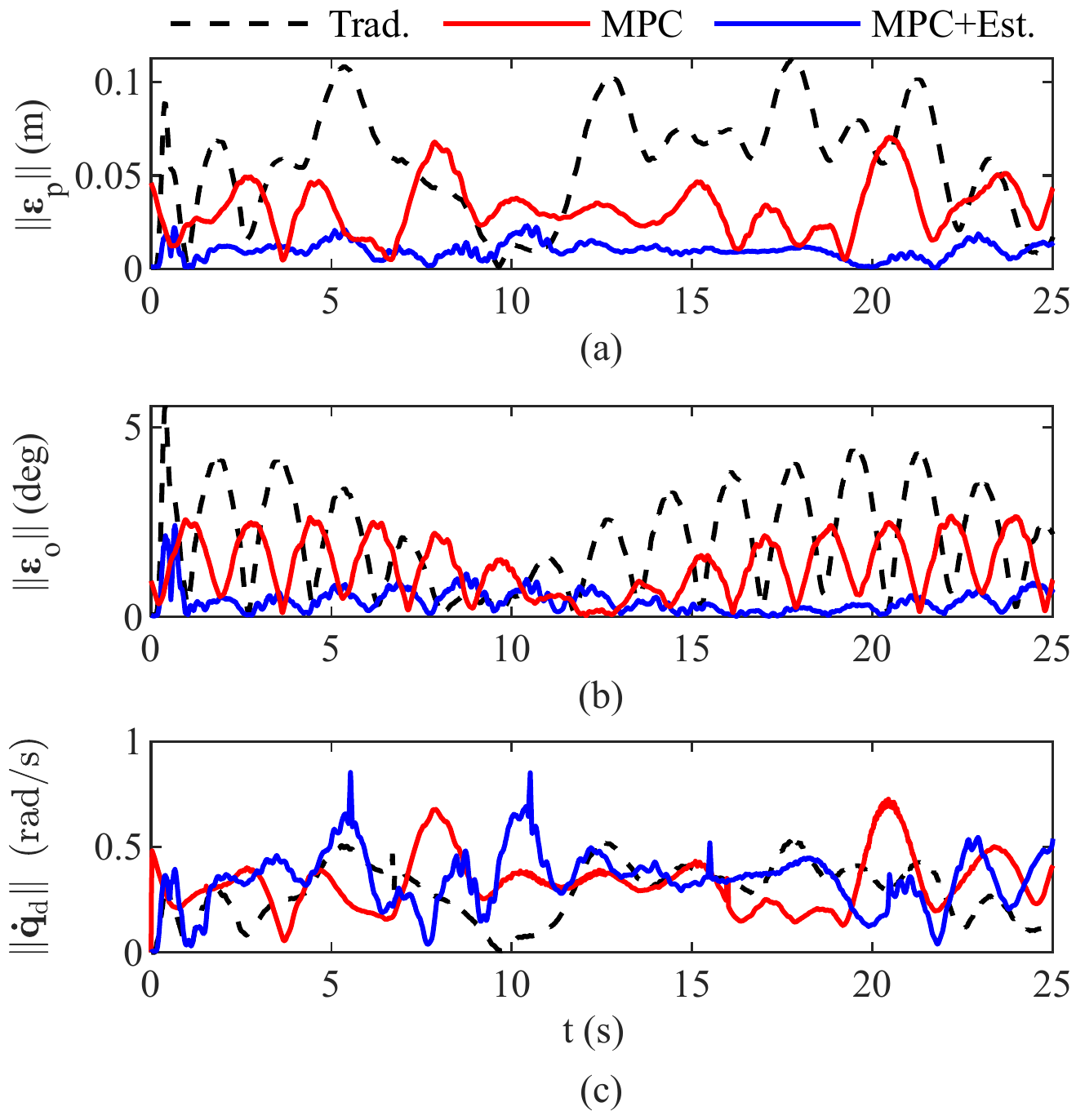}
	\caption{The indoor experimental results of trajectory tracking.}
	\label{exp_id2}
\end{figure}

In the second experiment, the manipulator is controlled to track a semi-arc trajectory whose radius is 0.47 m and the angular velocity is 0.15 rad. The results are shown in Fig. \ref{exp_id_traj}(b), red full line is the desired trajectory, and the result with estimator (blue full line) has higher accuracy than the result with MPC only (red dashed line). In Fig. \ref{exp_id2}(a), the average position error of a traditional controller is  0.057 m, while the maximum error reaches 0.113 m. The average error of the MPC controller is 0.032 m and the maximum error is 0.071 m. After the estimator is applied, the average error is reduced to 0.010 m, and the maximum position error is 0.023 m. In Fig. \ref{exp_id2}(b), the average orientation error of the traditional controller is 2.057 degrees, and the maximum error is 5.554 degrees. The average orientation error is reduced to 1.313 after the MPC controller is applied and the maximum error is reduced to 2.658 degrees. The average error is further reduced after the estimator is applied, which is 0.422 degrees, and the maximum error is 2.412 degrees. The outputs of three task controllers are given in Fig. \ref{exp_id2}(c), which reveals a similar conclusion as we introduced through Fig. \ref{exp_id}(c). The difference is that several singularities occur after the estimator is applied. This could be caused by the noises induced by the derivative of the tracking error in \eqref{eq3_13b}, and the square of the derivative amplifies the effect of noises.

\subsection{Field Experiments}
\begin{figure*}[t]
	\centering
	\setlength{\abovecaptionskip}{0pt}
	\setlength{\belowcaptionskip}{0pt}
	\includegraphics[width=1\linewidth]{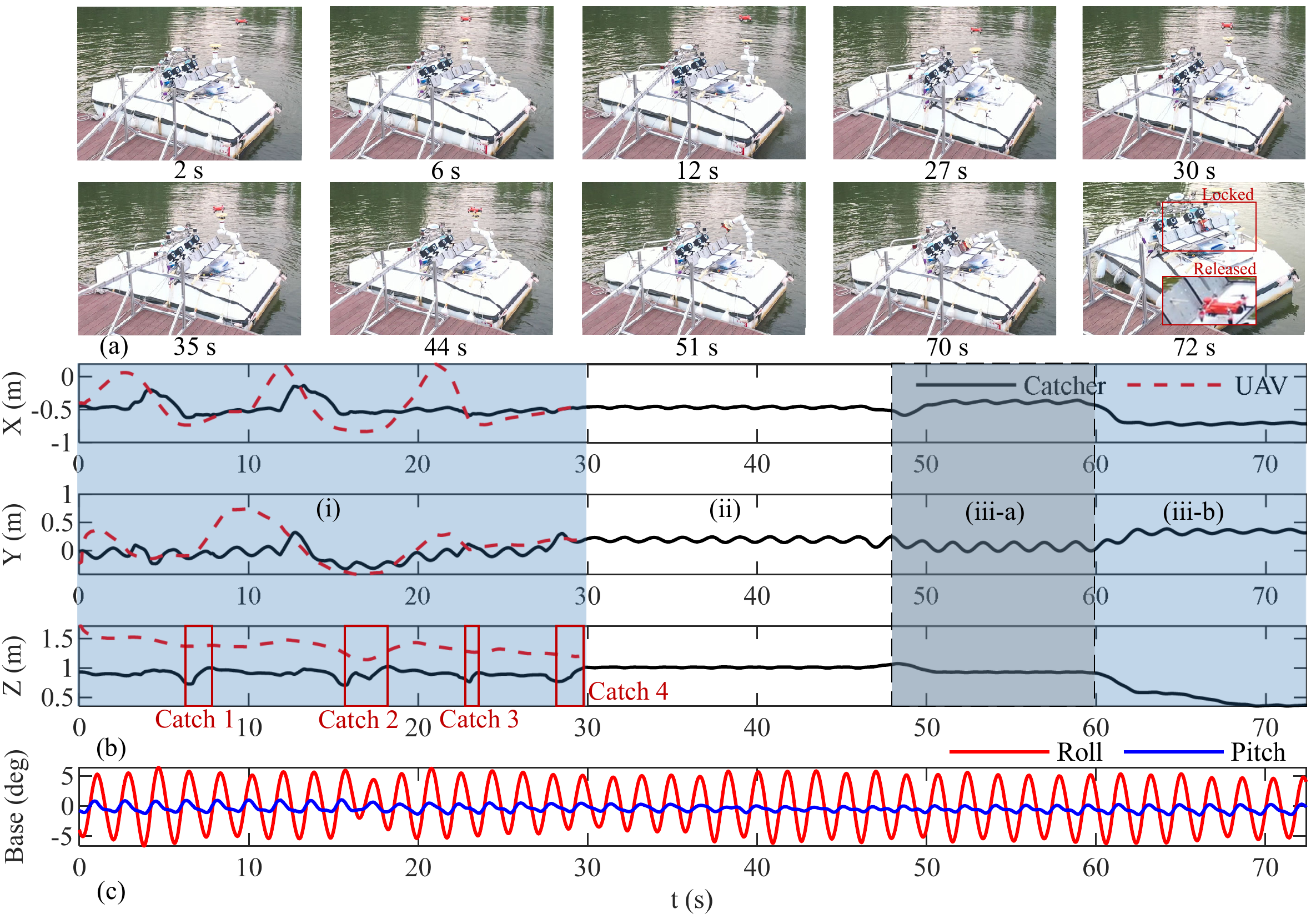}
	\caption{Field experiment results. (a) illustrates the working progress. (b) is the positions of the catcher and the UAV in the world frame. The roll and pitch of the USV are given in (c). The UAV is successfully caught after four tries of catching. (i)-(iii) are three phases of the landing assistance progress, which are the tracking and catching phase, tethered landing phase, and accurate placing phase. In phase (iii), (iii-a) is the UAV orientation calibration process, and (iii-b) is the process of placing the UAV on the landing platform.}
	\label{field_exp}
\end{figure*}
Field experiments are conducted to test the functionality of the landing assistance system. The vehicle-manipulator system is launched in Shenxian Lake, Shenzhen. We manually shook the USV to create a disturbing scenario, and an IMU is applied to sensor the pose of the USV. The motion of the USV is shown in Fig. \ref{field_exp} (c), the amplitude of the roll angle is about 6 degrees, and the period is about 1.67 seconds. A wireless station is established to communicate with the system. The onboard controller and offshore controller are connected to the same local area network, so we can send commands and collect data from the lakeshore. The four cameras in the camera system are calibrated in advance. The transformation between the manipulator base and the camera system is also calibrated. We mounted several marks on the end-effector of the manipulator, so the camera can capture the pose of the manipulator in the camera frame. Thereafter, the hand-eye calibration method can be applied to get the transformation between the camera and manipulator base.  

The experimental results are shown in Fig. \ref{field_exp}. The positions of the catcher and the UAV in the world frame are given in Fig. \ref{field_exp} (b), and the working progress with timestamp is illustrated in Fig. \ref{field_exp} (a). The three phases for landing assistance mentioned in Section \ref{SCS} are shown in (i)-(iii). in phase (i), the manipulator tracks the position of the UAV as shown in four figures captured at 2 seconds, 6 seconds, 12 seconds, and 27 seconds. The manipulator will catch the block hanging under the UAV as long as the catching condition mentioned in Section \ref{SCS} is satisfied. The end-effector position in the Z-direction shows that the manipulator tries four times and catches the block successfully at 30 seconds. The catcher locks the block at the moment the block is attracted. According to Fig. \ref{field_exp} (b), the catcher is not always kept tracking the UAV, and there are two main reasons. The first is that the UAV is out of the tracking space defined in \eqref{eq3_b_26}. The second reason is that the manipulator is in the process of restoration. The manipulator will restore to the initial position if one trial of catching is failed, and this process usually costs 4-6 seconds. In phase (ii), the manipulator captures the UAV by tethered landing. The UAV winds up the string to land on the catcher, as shown in the figure captured in 35 seconds. After the string is tightened, the UAV has a state shown in the figure at 44 seconds. After that, in phase (iii-a), the manipulator moves the UAV to the front of the camera, as shown in the figure captured at 51 seconds. The manipulator adjusts the UAV to a preset orientation. In phase (iii-b), The manipulator tracks a trajectory to place the UAV on the landing platform, as shown in, as shown in the figure at 70 seconds. Thereafter, the landing platform locks the UAV on the sloping plate, and the manipulator return to the initial position. At last, the platform pushes the UAV to the horizontal plate for takeoff.

\section{Conclusion}
In this work, a manipulator-based assistance system is developed to accurately and safely land the UAV on a disturbed USV. Field experiments are conducted with perpetuation imposed on the USV, and the results show that the UAV can be captured and placed on a small landing platform successfully. The manipulator controller is designed to track and catch the docking block hanging under the UAV, after which the end-effector is stabilized for the UAV tethered landing. An adaptive estimator is developed to improve the manipulator accuracy under the effect of the USV motion. Simulations and Indoor experiments are conducted to test the proposed controller. The results show that the position and orientation accuracy can be improved by 65$\%$ and 62$\%$ compared with an MPC controller, respectively, after the application of the estimator. The proposed assistance landing system combines the tethered landing and manipulator-assisted landing, making multiple UAVs able to land on one USV with a limited landing area.

\bibliographystyle{IEEEtran}
\bibliography{Vehicle_Manipulator.bib}

% Generated by IEEEtran.bst, version: 1.14 (2015/08/26)
\begin{thebibliography}{10}
\providecommand{\url}[1]{#1}
\csname url@samestyle\endcsname
\providecommand{\newblock}{\relax}
\providecommand{\bibinfo}[2]{#2}
\providecommand{\BIBentrySTDinterwordspacing}{\spaceskip=0pt\relax}
\providecommand{\BIBentryALTinterwordstretchfactor}{4}
\providecommand{\BIBentryALTinterwordspacing}{\spaceskip=\fontdimen2\font plus
\BIBentryALTinterwordstretchfactor\fontdimen3\font minus
  \fontdimen4\font\relax}
\providecommand{\BIBforeignlanguage}[2]{{%
\expandafter\ifx\csname l@#1\endcsname\relax
\typeout{** WARNING: IEEEtran.bst: No hyphenation pattern has been}%
\typeout{** loaded for the language `#1'. Using the pattern for}%
\typeout{** the default language instead.}%
\else
\language=\csname l@#1\endcsname
\fi
#2}}
\providecommand{\BIBdecl}{\relax}
\BIBdecl

\bibitem{zhang2020marine}
H.~Zhang, Y.~He, D.~Li, F.~Gu, Q.~Li, M.~Zhang, C.~Di, L.~Chu, B.~Chen, and
  Y.~Hu, ``Marine uav--usv marsupial platform: System and recovery technic
  verification,'' \emph{Applied Sciences}, vol.~10, no.~5, p. 1583, 2020.

\bibitem{lv2021analysis}
Z.~Lv, Y.~Bai, J.~Jin, H.~Wang, and C.~Ren, ``Analysis of wave fluctuation on
  underwater acoustic communication based usv,'' \emph{Applied Acoustics}, vol.
  175, p. 107820, 2021.

\bibitem{peng2017development}
Y.~Peng, Y.~Yang, J.~Cui, X.~Li, H.~Pu, J.~Gu, S.~Xie, and J.~Luo,
  ``Development of the usv ‘jinghai-i’and sea trials in the southern yellow
  sea,'' \emph{Ocean engineering}, vol. 131, pp. 186--196, 2017.

\bibitem{matus2021ground}
A.~Matus-Vargas, G.~Rodriguez-Gomez, and J.~Martinez-Carranza, ``Ground effect
  on rotorcraft unmanned aerial vehicles: a review,'' \emph{Intelligent Service
  Robotics}, vol.~14, no.~1, pp. 99--118, 2021.

\bibitem{maier2020coordinated}
M.~Maier, ``Coordinated control for robot-assisted take-off and landing of
  flying robots,'' Ph.D. dissertation, Technische Universit{\"a}t M{\"u}nchen,
  2020.

\bibitem{xu2020vision}
Z.-C. Xu, B.-B. Hu, B.~Liu, X.~Wang, and H.-T. Zhang, ``Vision-based autonomous
  landing of unmanned aerial vehicle on a motional unmanned surface vessel,''
  in \emph{2020 39th Chinese Control Conference (CCC)}.\hskip 1em plus 0.5em
  minus 0.4em\relax IEEE, 2020, pp. 6845--6850.

\bibitem{ambroziak2022experimental}
L.~Ambroziak, M.~Ciezkowski, A.~Wolniakowski, S.~Romaniuk, A.~Bozko,
  D.~Oldziej, and C.~Kownacki, ``Experimental tests of hybrid vtol unmanned
  aerial vehicle designed for surveillance missions and operations in maritime
  conditions from ship-based helipads,'' \emph{Journal of Field Robotics},
  vol.~39, no.~3, pp. 203--217, 2022.

\bibitem{borreguero2018experimental}
D.~Borreguero, O.~Velasco, and J.~Valente, ``Experimental design of a mobile
  landing platform to assist aerial surveys in fluvial environments,''
  \emph{Applied Sciences}, vol.~9, no.~1, p.~38, 2018.

\bibitem{lin2021low}
J.~Lin, Y.~Wang, Z.~Miao, H.~Zhong, and R.~Fierro, ``Low-complexity control for
  vision-based landing of quadrotor uav on unknown moving platform,''
  \emph{IEEE Transactions on Industrial Informatics}, vol.~18, no.~8, pp.
  5348--5358, 2021.

\bibitem{grlj2022decade}
C.~G. Grlj, N.~Krznar, and M.~Pranji{\'c}, ``A decade of uav docking stations:
  A brief overview of mobile and fixed landing platforms,'' \emph{Drones},
  vol.~6, no.~1, p.~17, 2022.

\bibitem{araar2017vision}
O.~Araar, N.~Aouf, and I.~Vitanov, ``Vision based autonomous landing of
  multirotor uav on moving platform,'' \emph{Journal of Intelligent \& Robotic
  Systems}, vol.~85, no.~2, pp. 369--384, 2017.

\bibitem{huang2021linear}
Y.~Huang, M.~Zhu, Z.~Zheng, and K.~H. Low, ``Linear velocity-free visual
  servoing control for unmanned helicopter landing on a ship with visibility
  constraint,'' \emph{IEEE Transactions on Systems, Man, and Cybernetics:
  Systems}, vol.~52, no.~5, pp. 2979--2993, 2021.

\bibitem{polvara2018vision}
R.~Polvara, S.~Sharma, J.~Wan, A.~Manning, and R.~Sutton, ``Vision-based
  autonomous landing of a quadrotor on the perturbed deck of an unmanned
  surface vehicle,'' \emph{drones}, vol.~2, no.~2, p.~15, 2018.

\bibitem{guo2022autonomous}
K.~Guo, P.~Tang, H.~Wang, D.~Lin, and X.~Cui, ``Autonomous landing of a
  quadrotor on a moving platform via model predictive control,''
  \emph{Aerospace}, vol.~9, no.~1, p.~34, 2022.

\bibitem{paris2020dynamic}
A.~Paris, B.~T. Lopez, and J.~P. How, ``Dynamic landing of an autonomous
  quadrotor on a moving platform in turbulent wind conditions,'' in \emph{2020
  IEEE International Conference on Robotics and Automation (ICRA)}.\hskip 1em
  plus 0.5em minus 0.4em\relax IEEE, 2020, pp. 9577--9583.

\bibitem{baca2019autonomous}
T.~Baca, P.~Stepan, V.~Spurny, D.~Hert, R.~Penicka, M.~Saska, J.~Thomas,
  G.~Loianno, and V.~Kumar, ``Autonomous landing on a moving vehicle with an
  unmanned aerial vehicle,'' \emph{Journal of Field Robotics}, vol.~36, no.~5,
  pp. 874--891, 2019.

\bibitem{sinisterra2017usv}
A.~Sinisterra, M.~Dhanak, and N.~Kouvaras, ``A usv platform for surface
  autonomy,'' in \emph{OCEANS 2017-Anchorage}.\hskip 1em plus 0.5em minus
  0.4em\relax IEEE, 2017, pp. 1--8.

\bibitem{ross2022autonomous}
J.~Ross, M.~Seto, and C.~Johnston, ``Autonomous landing of rotary wing unmanned
  aerial vehicles on underway ships in a sea state,'' \emph{Journal of
  Intelligent \& Robotic Systems}, vol. 104, no.~1, pp. 1--9, 2022.

\bibitem{oh2006approaches}
S.-R. Oh, K.~Pathak, S.~K. Agrawal, H.~R. Pota, and M.~Garratt, ``Approaches
  for a tether-guided landing of an autonomous helicopter,'' \emph{IEEE
  Transactions on Robotics}, vol.~22, no.~3, pp. 536--544, 2006.

\bibitem{kourani2021marine}
A.~Kourani and N.~Daher, ``Marine locomotion: A tethered uav-buoy system with
  surge velocity control,'' \emph{Robotics and Autonomous Systems}, vol. 145,
  p. 103858, 2021.

\bibitem{kishk20203}
M.~A. Kishk, A.~Bader, and M.-S. Alouini, ``On the 3-d placement of airborne
  base stations using tethered uavs,'' \emph{IEEE Transactions on
  Communications}, vol.~68, no.~8, pp. 5202--5215, 2020.

\bibitem{liu2022design}
C.~Liu, Z.~Jiang, R.~Xu, X.~Ji, L.~Zhang, and H.~Qian, ``Design and
  optimization of a magnetic catcher for uav landing on disturbed aquatic
  surface platforms,'' in \emph{2022 International Conference on Robotics and
  Automation (ICRA)}.\hskip 1em plus 0.5em minus 0.4em\relax IEEE, 2022, pp.
  1162--1168.

\bibitem{kominami2021active}
T.~Kominami, H.~Paul, R.~Miyazaki, B.~Sumetheeprasit, R.~Ladig, and
  K.~Shimonomura, ``Active tethered hook: Heavy load movement using hooks that
  move actively with micro uavs and winch system,'' in \emph{2021 IEEE/ASME
  International Conference on Advanced Intelligent Mechatronics (AIM)}.\hskip
  1em plus 0.5em minus 0.4em\relax IEEE, 2021, pp. 264--269.

\bibitem{schuchardt2021maritime}
B.~I. Schuchardt, T.~Dautermann, A.~Donkels, S.~Krause, N.~Peinecke, and
  G.~Schwoch, ``Maritime operation of an unmanned rotorcraft with tethered ship
  deck landing system,'' \emph{CEAS Aeronautical Journal}, vol.~12, no.~1, pp.
  3--11, 2021.

\bibitem{nguyen2019passively}
H.-N. Nguyen, R.~Siddall, B.~Stephens, A.~Navarro-Rubio, and M.~Kova{\v{c}},
  ``A passively adaptive microspine grapple for robust, controllable
  perching,'' in \emph{2019 2nd IEEE International Conference on Soft Robotics
  (RoboSoft)}.\hskip 1em plus 0.5em minus 0.4em\relax IEEE, 2019, pp. 80--87.

\bibitem{maier2015robot}
M.~Maier, A.~Oeschger, and K.~Kondak, ``Robot-assisted landing of vtol uavs:
  Design and comparison of coupled and decoupling linear state-space control
  approaches,'' \emph{IEEE Robotics and Automation Letters}, vol.~1, no.~1, pp.
  114--121, 2015.

\bibitem{from2010singularity}
P.~J. From, V.~Duindam, K.~Y. Pettersen, J.~T. Gravdahl, and S.~Sastry,
  ``Singularity-free dynamic equations of vehicle--manipulator systems,''
  \emph{Simulation Modelling Practice and Theory}, vol.~18, no.~6, pp.
  712--731, 2010.

\bibitem{from2011motion}
P.~J. From, J.~T. Gravdahl, T.~Lillehagen, and P.~Abbeel, ``Motion planning and
  control of robotic manipulators on seaborne platforms,'' \emph{Control
  engineering practice}, vol.~19, no.~8, pp. 809--819, 2011.

\bibitem{kosuge1991control}
K.~Kosuge, T.~Fukuda, and H.~Ohkubo, ``Control of manipulator/vehicle system
  floating on the water,'' in \emph{[1991] Proceedings of the 30th IEEE
  Conference on Decision and Control}.\hskip 1em plus 0.5em minus 0.4em\relax
  IEEE, 1991, pp. 2781--2786.

\bibitem{xu2020novel}
R.~Xu, H.~Liu, C.~Liu, Z.~Sun, T.~L. Lam, and H.~Qian, ``A novel solar tracker
  driven by waves: From idea to implementation,'' in \emph{2020 IEEE
  International Conference on Robotics and Automation (ICRA)}.\hskip 1em plus
  0.5em minus 0.4em\relax IEEE, 2020, pp. 8209--8214.

\bibitem{yuguang2019dynamic}
Y.~Zhong and F.~Yang, ``Dynamic modeling and adaptive fuzzy sliding mode
  control for multi-link underwater manipulators,'' \emph{Ocean Engineering},
  vol. 187, p. 106202, 2019.

\bibitem{dai2022fast}
Y.~Dai, H.~Gao, S.~Yu, and Z.~Ju, ``A fast tube model predictive control scheme
  based on sliding mode control for underwater vehicle-manipulator system,''
  \emph{Ocean Engineering}, vol. 254, p. 111259, 2022.

\bibitem{zhang2021manipulator}
F.~Zhang and G.~Duan, ``Manipulator actuated integrated position and attitude
  stabilization of spacecraft subject to external disturbances,'' \emph{IEEE
  Transactions on Systems, Man, and Cybernetics: Systems}, 2021.

\bibitem{guerrero2019observation}
J.~Guerrero, J.~Torres, V.~Creuze, and A.~Chemori, ``Observation-based
  nonlinear proportional--derivative control for robust trajectory tracking for
  autonomous underwater vehicles,'' \emph{IEEE journal of oceanic engineering},
  vol.~45, no.~4, pp. 1190--1202, 2019.

\bibitem{qiang2020model}
H.~Qiang, S.~Jin, X.~Feng, D.~Xue, and L.~Zhang, ``Model predictive control of
  a shipborne hydraulic parallel stabilized platform based on ship motion
  prediction,'' \emph{IEEE Access}, vol.~8, pp. 181\,880--181\,892, 2020.

\bibitem{woolfrey2021predictive}
J.~Woolfrey, W.~Lu, and D.~Liu, ``Predictive end-effector control of
  manipulators on moving platforms under disturbance,'' \emph{IEEE Transactions
  on Robotics}, vol.~37, no.~6, pp. 2210--2217, 2021.

\bibitem{xu2021predictive}
R.~Xu, X.~Ji, J.~Hou, H.~Liu, and H.~Qian, ``A predictive control method for
  stabilizing a manipulator-based uav landing platform on fluctuating marine
  surface,'' in \emph{2021 IEEE/RSJ International Conference on Intelligent
  Robots and Systems (IROS)}.\hskip 1em plus 0.5em minus 0.4em\relax IEEE,
  2021, pp. 8625--8632.

\bibitem{qian2021adaptive}
Y.~Qian, D.~Hu, Y.~Chen, Y.~Fang, and Y.~Hu, ``Adaptive neural network-based
  tracking control of underactuated offshore ship-to-ship crane systems subject
  to unknown wave motions disturbances,'' \emph{IEEE Transactions on systems,
  man, and cybernetics: systems}, vol.~52, no.~6, pp. 3626--3637, 2021.

\bibitem{han2021adaptive}
L.~Han, G.~Tang, M.~Cheng, H.~Huang, and D.~Xie, ``Adaptive nonsingular fast
  terminal sliding mode tracking control for an underwater vehicle-manipulator
  system with extended state observer,'' \emph{Journal of Marine Science and
  Engineering}, vol.~9, no.~5, p. 501, 2021.

\bibitem{xu2022design}
R.~Xu, X.~Ji, C.~Liu, J.~Hou, Z.~Cao, and H.~Qian, ``Design and control of a
  wave-driven solar tracker,'' \emph{IEEE Transactions on Automation Science
  and Engineering}, 2022.

\bibitem{leahy1989compensation}
M.~B. Leahy~Jr and G.~N. Saridis, ``Compensation of industrial manipulator
  dynamics,'' \emph{The International Journal of Robotics Research}, vol.~8,
  no.~4, pp. 73--84, 1989.

\bibitem{ahmadi2016robust}
S.~M. Ahmadi and M.~M. Fateh, ``Robust control of electrically driven robots
  using adaptive uncertainty estimation,'' \emph{Computers \& Electrical
  Engineering}, vol.~56, pp. 674--687, 2016.

\bibitem{wahrburg2016mpc}
A.~Wahrburg and K.~Listmann, ``Mpc-based admittance control for robotic
  manipulators,'' in \emph{2016 IEEE 55th Conference on Decision and Control
  (CDC)}.\hskip 1em plus 0.5em minus 0.4em\relax IEEE, 2016, pp. 7548--7554.

\bibitem{flacco2015control}
F.~Flacco, A.~De~Luca, and O.~Khatib, ``Control of redundant robots under hard
  joint constraints: Saturation in the null space,'' \emph{IEEE Transactions on
  Robotics}, vol.~31, no.~3, pp. 637--654, 2015.

\bibitem{osqp}
\BIBentryALTinterwordspacing
B.~Stellato, G.~Banjac, P.~Goulart, A.~Bemporad, and S.~Boyd, ``{OSQP}: an
  operator splitting solver for quadratic programs,'' \emph{Mathematical
  Programming Computation}, vol.~12, no.~4, pp. 637--672, 2020. [Online].
  Available: \url{https://doi.org/10.1007/s12532-020-00179-2}
\BIBentrySTDinterwordspacing

\bibitem{kai2009robust}
X.~Kai, C.~Wei, and L.~Liu, ``Robust extended kalman filtering for nonlinear
  systems with stochastic uncertainties,'' \emph{IEEE Transactions on Systems,
  Man, and Cybernetics-Part A: Systems and Humans}, vol.~40, no.~2, pp.
  399--405, 2009.

\end{thebibliography}

\vfill

\end{document}